\documentclass[sigconf]{acmart}

\usepackage{booktabs} 
\usepackage{multirow}
\usepackage{subfloat}
\usepackage{caption}
\usepackage{subfig}
\usepackage{pgfplots}
\usepackage{tikz}

\DeclareMathOperator{\val}{=}  

\def\happensAt{\textsf{\small happensAt}}

\def\holdsAt{\textsf{\small holdsAt}}
\def\holdsFor{\textsf{\small holdsFor}}

\def\initiatedAt{\textsf{\small initiatedAt}}
\def\terminatedAt{\textsf{\small terminatedAt}}

\def\startE{\textsf{\small start}}
\def\endE{\textsf{\small end}}

\def\unionall{\textsf{\small union\_all}}

\def\intersectall{\textsf{\small intersect\_all}}

\def\complementall{\textsf{\small relative\_complement\_all}}

\def\nbf{\textsf{\small not}}
\def\true{\textsf{\small true}}

\newenvironment{mysplit}%
  {\arraycolsep 0pt \begin{array}{l}}%
  {\end{array}}
  
\newcommand\arraybslash{\let\\\@arraycr}
\renewcommand\arraystretch{1.3}
\newcounter{Table}

\renewcommand\footnotetextcopyrightpermission[1]{} 
\setcopyright{none}

\begin{document}
\title{Composite Event Recognition for Maritime Monitoring}

\author{Manolis Pitsikalis$^{1}$, Alexander Artikis$^{2,1}$, 
 Richard Dreo$^{3}$,Cyril Ray$^{3,4}$, Elena Camossi$^{5}$ and Anne-Laure Jousselme$^{5}$}
\affiliation{
  \institution{$^{1}$Institute of Informatics \& Telecommunications, NCSR Demokritos, Athens, Greece}
    \institution{$^2$Department of Maritime Studies, University of Piraeus, Greece}
    \institution{$^3$Naval Academy Research Institute, Brest, France, $^4$Arts et Metiers ParisTech, France}
    \institution{$^{5}$Centre for Maritime Research and Experimentation (CMRE), NATO, La Spezia, Italy}
}
\email{{manospits, a.artikis}@iit.demokritos.gr, {richard.dreo, cyril.ray}@ecole-navale.fr,} 
\email{{elena.camossi, anne-laure.jousselme}@cmre.nato.int}

\renewcommand{\shortauthors}{}

\begin{abstract}
Maritime monitoring systems support safe shipping as they allow for the real-time detection of dangerous, suspicious and illegal vessel activities. We present such a system using the Run-Time Event Calculus, a composite event recognition system with formal, declarative semantics. For effective recognition, we developed a library of maritime patterns in close collaboration with domain experts. We present a thorough evaluation of the system and the patterns both in terms of predictive accuracy and computational efficiency, using real-world datasets of vessel position streams and contextual geographical information. 
\end{abstract}



\maketitle

\section{Introduction}

The International Maritime Organization (IMO)\footnote{\url{http://www.imo.org/en/OurWork/Environment/Pages/Default.aspx}} states that 90\% of the global trade is handled by the shipping industry. 
Maritime monitoring systems support safe shipping by detecting, in real-time, dangerous, suspicious and illegal vessel activities. Such systems typically use the Automatic Identification System (AIS)\footnote{\url{http://www.imo.org/en/OurWork/Safety/Navigation/Pages/AIS.aspx}}, a tracking technology for locating vessels at sea through data exchange. AIS integrates a VHF transceiver with a positioning device (e.g., GPS), and other electronic navigation sensors, such as a gyrocompass or rate of turn indicator, thus producing valuable data regarding the vessel and its current status. The acquisition of positional data is achieved by AIS base stations along coastlines, or even by satellites when out of range of terrestrial networks. 

Maritime monitoring systems have been attracting considerable attention for economic as well as environmental reasons \cite{jousselme_anne_laure_2016_804845, 6384120, IdiriNapoli12, Snidaro:2015:FUK:2657516.2657960, Terroso-Saenz2016, Riveiro428946, datacron-del5.1, 7363883, rs9030246}. Terosso-Saenz et al.~\cite{Terroso-Saenz2016}, e.g., presented a system detecting abnormally high or low vessel speed, as well as when two vessels are in danger of colliding. 
SUMO \cite{rs9030246} is an open-source system combining AIS streams with synthetic aperture radar images for detecting illegal oil dumping, piracy and unsustainable fishing. 
van Laere et al.~\cite{Laere09} evaluated a workshop aiming at the identification of potential vessel anomalies, such as tampering, rendez-vous between vessels and unusual routing.   
 
In previous work, we presented a maritime monitoring system~\cite{DBLP:journals/geoinformatica/PatroumpasAAVPT17} with a component for trajectory simplification, allowing for efficient maritime stream analytics, and a composite event recognition component, combining kinematic vessel streams with contextual (geographical) knowledge for real-time vessel activity detection. To improve the accuracy of the system, we collaborated, in the context of this paper, with domain experts in order to construct effective patterns of maritime activity. Thus, we present a library of such patterns in the Event Calculus \cite{DBLP:journals/tkde/ArtikisSP15}---the language of the composite event recognition component. Furthermore, we evaluate these patterns and the maritime monitoring system both in terms of predictive accuracy and efficiency, using real datasets. 

Our data comes from the datAcron project\footnote{\url{http://datacron-project.eu/}}. First, we used a publicly available stream of 18M AIS position signals, transmitted by 5K vessels sailing in the Atlantic Ocean around the port of Brest, France, between October 2015 and March 2016~\cite{RayData}. Second, we employed a stream of 55M terrestrial and satellite AIS position signals transmitted by 34K vessels during January 2016 in the European seas \cite{datacron-del5.3}. These streams are consumed in conjunction with static geographical data, such as areas represented as polygons---fishing areas~\cite{doi:10.1080/17445647.2016.1195299,10.1371/journal.pone.0130746}, (protected) Natura 2000 areas\footnote{\url{ http://ec.europa.eu/environment/nature/natura2000/index_en.htm}}, etc.

The contributions of this paper are then the following:
\begin{itemize}
 \item A  monitoring system with a formal specification of effective maritime patterns. In contrast to related systems, these patterns concern a wide range of maritime activities. 
 \item An evaluation of the system and the patterns in terms of both efficiency and accuracy using large, real datasets.
\end{itemize}

All composite maritime activities that were recognised on the dataset of Brest have been made publicly available \cite{Mar-CER-dataset}, in order to aid further research, such as the development of machine learning algorithms for pattern construction and classification.

The remainder of the paper is structured as follows. In Section \ref{sec:background} we present the background of this work. 
Then, in Section~\ref{sec:maritime} we present the maritime patterns in the language of RTEC, our  composite event recognition engine. The empirical evaluation is presented in Section \ref{sec:analysis}, while in Section \ref{sec:summary} we summarise the work and discuss further research directions.

\section{Background}
\label{sec:background}

\subsection{Spatial Preprocessing \& Trajectory Simplification}

\begin{figure}[t]
    \centering
    \includegraphics[width=\columnwidth]{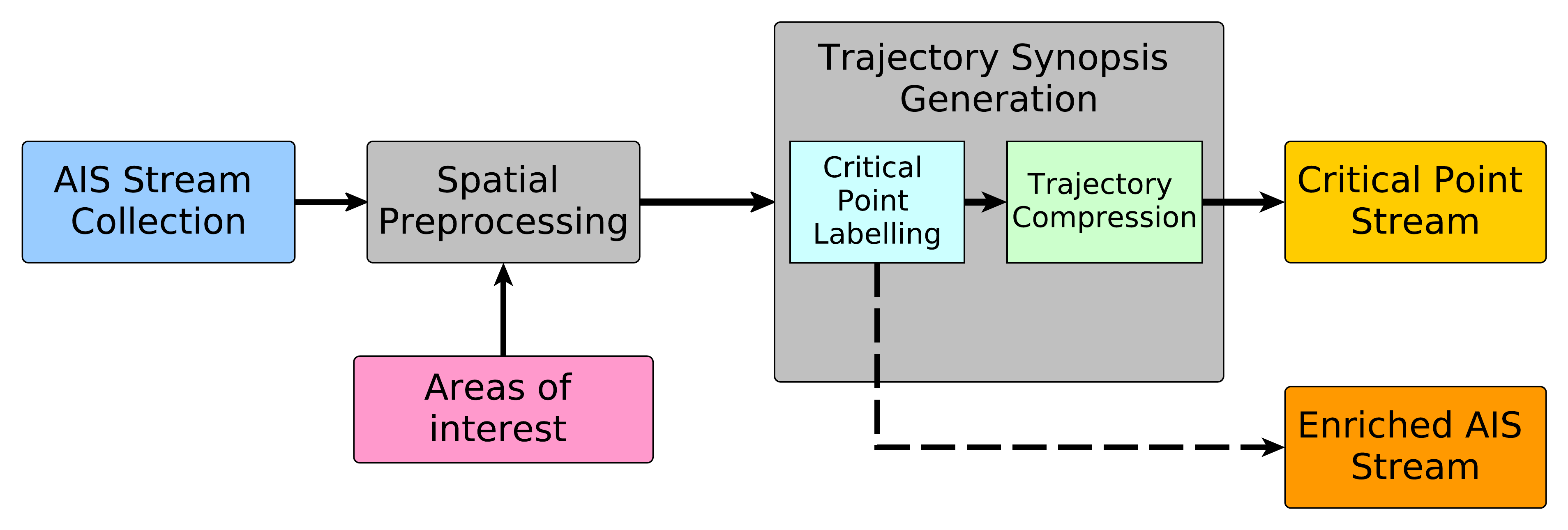}
    \caption{Steps prior to Composite Event Recognition.}
    \label{fig:prior-CER}
\end{figure}

Composite event recognition (CER) on vessel position signals, as we have defined it \cite{DBLP:journals/geoinformatica/PatroumpasAAVPT17}, requires two online tasks/steps: (a) computing a set of spatial relations among vessels, such as proximity, and among vessels and areas of interest (e.g., fishing areas), and (b) labelling position signals of interest as `critical'---such as when a vessel changes its speed, turns, stops, moves slowly or stops transmitting its position. Figure \ref{fig:prior-CER} illustrates these steps. AIS position signals are streamed into the system, and go through a spatial preprocessing step, for the computation of the spatial relations required by the composite event patterns \cite{DBLP:conf/time/SantipantakisVD18}. These relations are displayed at the top of Table \ref{tbl:events}.
Then, the relevant position signals are annotated as critical---see the middle part of Table \ref{tbl:events}. Subsequently, the position signals may be consumed by the CER component either directly (see `enriched AIS stream' in Figure \ref{fig:prior-CER}), or after being compressed, i.e.~after removing all signals that have not been labelled as critical (see `critical point stream' in Figure \ref{fig:prior-CER}). In Section \ref{sec:analysis}, we present the effects of such a compression on the efficiency and the accuracy of the system.

Critical point labelling is performed as part of trajectory synopsis generation, whereby  major changes along each vessel's movement are tracked. This process can instantly identify critical points along each trajectory, such as a stop, a turn, or slow motion. Using the retained critical points, we may reconstruct a vessel trajectory with small acceptable deviations from the original one. Empirical results have indicated that 70-80\% of the input data may be discarded as redundant, while compression ratio can be up to 99\% when the frequency of position updates is high \cite{DBLP:journals/geoinformatica/PatroumpasAAVPT17}. 

\subsection{Composite Event Recognition}
    \label{sec:RTEC}
    
We perform CER using the `Event Calculus for Run-Time reasoning' (RTEC) \cite{DBLP:journals/tkde/ArtikisSP15,DBLP:journals/igpl/ArtikisS10}, an open-source Prolog implementation of the Event Calculus \cite{DBLP:journals/ngc/KowalskiS86}, designed to compute continuous narrative assimilation queries for pattern matching on data streams.
RTEC has a formal, declarative semantics---composite patterns
are (locally) stratified logic programs \cite{local-strat}. Moreover, RTEC includes optimisation techniques for efficient pattern matching, such as `windowing', whereby all input events that took place prior to the current window  are discarded/`forgotten'. Details about the reasoning algorithms of RTEC, including a complexity analysis, may be found in~\cite{DBLP:journals/tkde/ArtikisSP15}.

\begin{table}[t]
\caption{Composite Event/Activity Recognition: Input events are presented above the double horizontal line, while the output stream is presented below this line. The input events above the single horizontal line are detected at the spatial preprocessing step while the remaining ones are detected by the trajectory synopsis generator (critical events). 
With the exception of $\mathit{proximity}$, all items of the input stream are instantaneous, while all output activities are durative.
}\label{tbl:events}\vspace{-0.4cm}
\begin{center}
\renewcommand{\arraystretch}{0.9}
\setlength\tabcolsep{1pt}
\begin{tabular}{lll}
\hline\noalign{\smallskip}
\multicolumn{1}{c}{}& \multicolumn{1}{l}{\textbf{Event/Activity}} & \multicolumn{1}{c}{\textbf{Description}}  \\
\noalign{\smallskip}
\hline
\noalign{\smallskip}
\multirow{3}{*}{\rotatebox[origin=c]{90}{\textbf{Spatial}}}    
&$\mathit{entersArea(V,A)}$ & Vessel $V$ enters area $A$ \\[2pt]
&$\mathit{leavesArea(V,A)}$ & Vessel $V$ leaves area $A$\\[2pt]
&$\mathit{proximity(V_1,V_2)}$ & Vessels $V_1$ and $V_2$ are close to each other\\
\noalign{\smallskip}
\hline
\noalign{\smallskip}
\multirow{13}{*}{\rotatebox[origin=c]{90}{\textbf{Critical}}}    
&$\mathit{gap\_start(V)}$ & Vessel $V$ stopped sending \\
& & position signals \\[2pt]
&$\mathit{gap\_end(V)}$   & Vessel $V$ resumed sending \\
& & position signals \\[2pt]
&$\mathit{slow\_motion\_start(V)}$ & Vessel $V$ started moving at a low speed \\[2pt]
&$\mathit{slow\_motion\_end(V)}$ & Vessel $V$ stopped moving \\
& & at a low speed \\[2pt]
&$\mathit{stop\_start(V)}$ & Vessel $V$ started being idle \\[2pt]
&$\mathit{stop\_end(V)}$ & Vessel $V$ stopped being idle \\[2pt]
&$\mathit{change\_in\_}$ & Vessel $V$ started changing  its speed \\
&$\mathit{speed\_start(V)}$ & \\[2pt]
&$\mathit{change\_in\_}$ & Vessel $V$ stopped changing its speed \\
&$\mathit{speed\_end(V)}$ &  \\[2pt]
&$\mathit{change\_in\_heading(V)}$ & Vessel $V$ changed its heading \\
\noalign{\smallskip}
\hline\hline
\noalign{\smallskip}
\multirow{13}{*}{\rotatebox[origin=c]{90}{\textbf{Composite}}}
&$\mathit{highSpeedNC(V)}$ & Vessel $V$ has high speed near coast\\[2pt]
&$\mathit{anchoredOrMoored(V)}$ & Vessel $V$ is anchored or moored\\[2pt]
&$\mathit{drifting(V)}$       & Vessel $V$ is drifting\\[2pt]
&$\mathit{trawling(V)}$ & Vessel $V$ is trawling\\[2pt]
&$\mathit{tugging(V_1,V_2)}$ & Vessels $V_1$ and $V_2$ are engaged \\
& & in tugging \\[2pt]
&$\mathit{pilotBoarding(V_1,V_2)}$ & Vessels $V_1$ and $V_2$ are engaged in \\
&& pilot boarding\\[2pt]
&$\mathit{rendezVous(V_1,V_2)}$ & Vessels $V_1$ and $V_2$ are having \\
& &  a rendez-vous \\[2pt]
&$\mathit{loitering(V)}$ & Vessel $V$ is loitering\\[2pt]
&$\mathit{sar(V)}$ & Vessel $V$ is engaged in a \\
&& search and rescue (SAR) operation\\[2pt]
\hline
\end{tabular}
\end{center}
\end{table}

\begin{table*}[t]
\caption{Main predicates of RTEC. `$F\val V$' denotes that fluent $F$ has value $V$.}\label{tbl:ec}\vspace{-0.4cm}
\begin{center}
\renewcommand{\arraystretch}{0.9}
\setlength\tabcolsep{3.6pt}
\begin{tabular}{ll}
\hline\noalign{\smallskip}
\multicolumn{1}{c}{\textbf{Predicate}} & \multicolumn{1}{c}{\textbf{Meaning}}  \\
\noalign{\smallskip}
\hline
\noalign{\smallskip}
\happensAt$(E, T)$ & Event $E$ occurs at time $T$  \\[2pt]
\holdsAt$\mathit{(F \val V, T)}$ & The value of fluent $F$ is $V$ at time $T$ \\[3pt]
\holdsFor$\mathit{(F \val V, I)}$ & $I$ is the list of the maximal intervals for which $F\val V$ holds continuously\\[2pt]
\initiatedAt$\mathit{(F \val V, T)}$ & At time $T$ a period of time for which $F\val V$ is initiated \\[2pt]
\terminatedAt$\mathit{(F \val V, T)}$ & At time $T$ a period of time for which $F\val V$ is terminated \\[2pt]
\unionall$\mathit{(L, I)}$ & $\mathit{I}$ is the list of maximal intervals produced by the union of the lists of maximal intervals of list $L$ \\[2pt]
\intersectall$\mathit{(L, I)}$ & $\mathit{I}$ is the list of maximal intervals produced by the intersection of the lists of maximal intervals of list $\mathit{L}$ \\[2pt]
\complementall$\mathit{(I', L, I)}$ & $I$ is the list of maximal intervals produced by the relative complement of the list\\& of maximal intervals $I'$ with respect to every list of maximal intervals of list $L$ \\[2pt]
\hline
\end{tabular}
\end{center}
\end{table*}
\begin{figure*}[t]
\renewcommand{\thesubfigure}{a}
\subfloat[Union.]{\centering
\includegraphics[width=.25\textwidth]{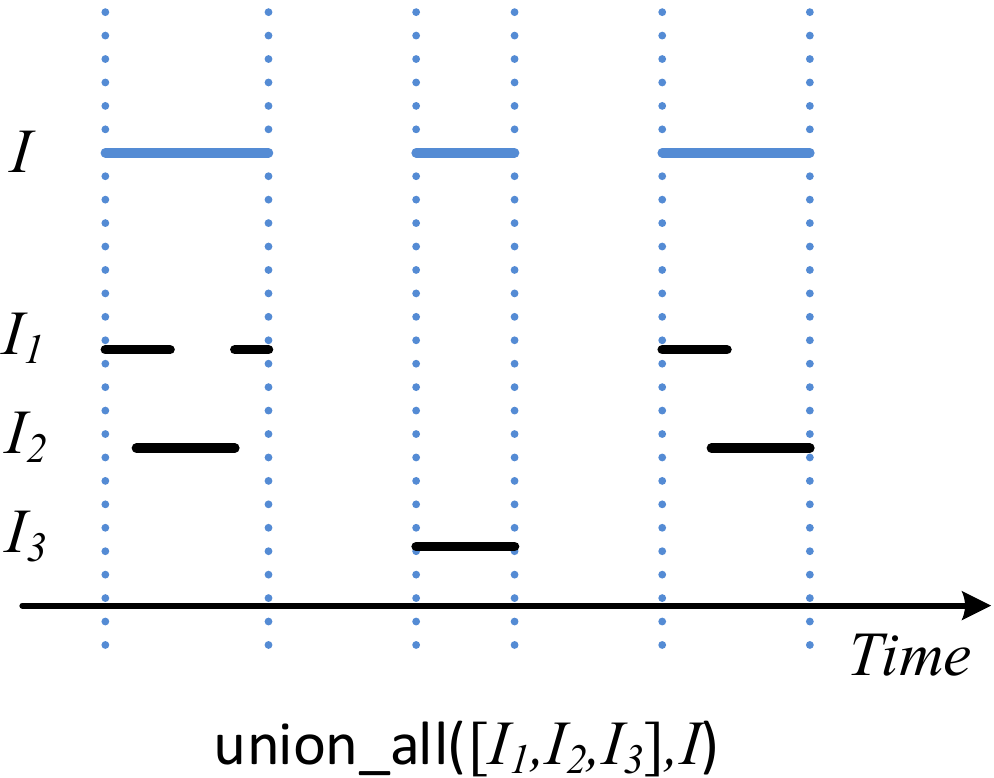}
\label{fig:union}}$\quad$
\renewcommand{\thesubfigure}{b}
\subfloat[Intersection.]{\centering
\includegraphics[width=.25\textwidth]{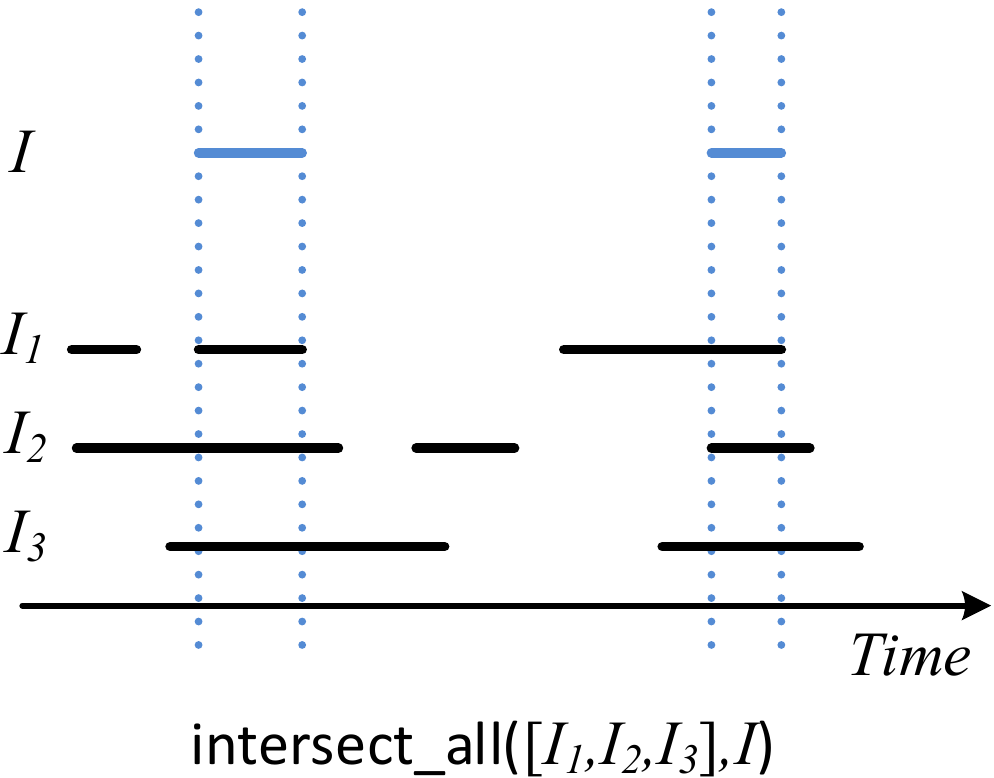}
\label{fig:intersection}}$\quad$
\renewcommand{\thesubfigure}{c}
\subfloat[Relative Complement.]{\centering
\includegraphics[width=.25\textwidth]{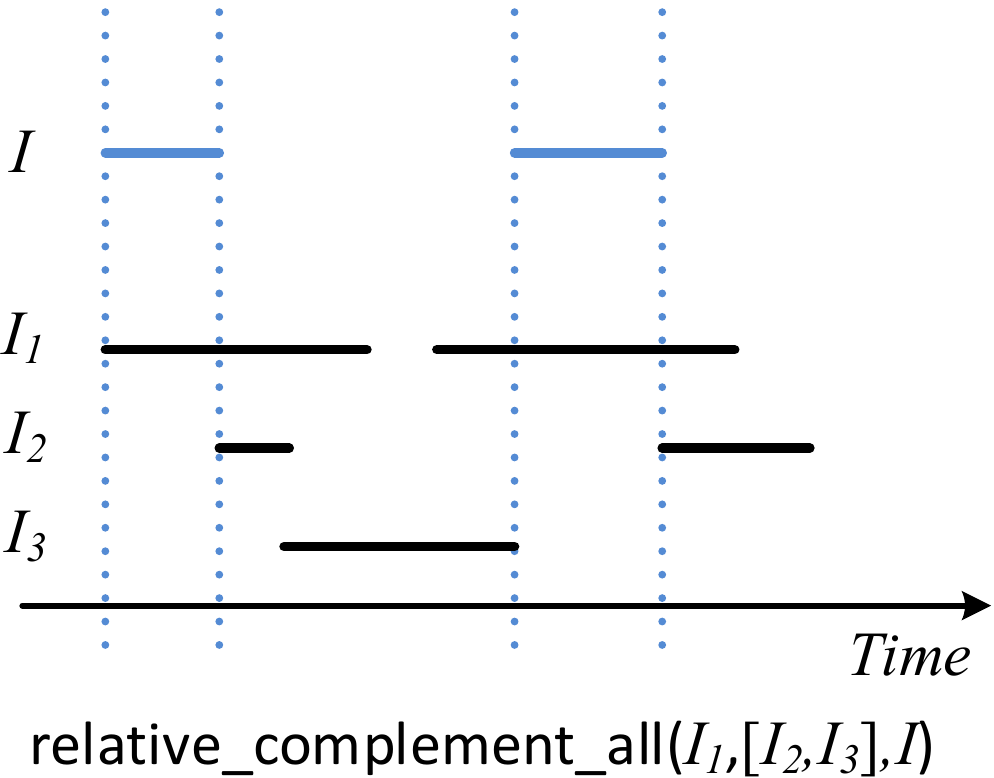}
\label{fig:rcomplement}}
\caption{A visual illustration of the interval manipulation constructs of RTEC. In these examples, there are three input streams, $I_1$, $I_2$ and $I_3$, coloured black. The output of each interval manipulation construct $I$ is coloured light blue.}
\label{fig:interval-manipulation}
\end{figure*}

The time model in RTEC is linear and includes integer time-points. 
An \emph{event description} includes rules that define the event instances with the use of the $\mathit{\happensAt}$ predicate,
the effects of events on \emph{fluents}---time-varying properties---with the use of the $\mathit{\initiatedAt}$ and $\mathit{\terminatedAt}$ predicates, and the values of the fluents with the use of the $\mathit{\holdsAt}$ and $\mathit{\holdsFor}$ predicates. Table \ref{tbl:ec} summarises the main predicates of RTEC. 

Fluents are `simple' or `statically determined'. In brief, simple fluents are defined by means of \initiatedAt\ and \terminatedAt\ rules, while statically determined fluents are defined by means of application-dependent \holdsFor\ rules, along with the interval manipulation constructs of RTEC: \unionall, \intersectall\ and \complementall. See Table \ref{tbl:ec} for a brief explanation of these constructs and Figure \ref{fig:interval-manipulation} for an example visualisation. Composite events/activities are typically durative; thus the task generally is to compute the maximal intervals for which a fluent
expressing a composite activity has a particular value continuously. Below, we discuss the representation of fluents/composite maritime activities, and briefly present the way in which we compute their maximal intervals.

\section{Maritime Activity Patterns}
\label{sec:maritime}

In previous work, we developed a system for maritime surveillance focusing mostly on efficiency \cite{DBLP:journals/geoinformatica/PatroumpasAAVPT17}. To improve the accuracy of the system, we collaborated with the domain experts of the datAcron project in order to construct effective patterns of maritime activity. In what follows, we present a formalisation of some of these patterns in the language of RTEC.

\subsection{Building Blocks}

We begin by presenting a set of building blocks that will be later used for the construction of more involved patterns.

\subsubsection{Vessel within area of interest}

Calculating the time periods during which a vessel is in an area of some type, such as a (protected) Natura 2000, fishing or anchorage area, is particularly useful in maritime (e.g.~fishing) patterns. Consider the formalisation below:
\begin{align} 
\begin{mysplit}
\label{eq:withinArea}
\initiatedAt\mathit{(withinArea(Vessel,AreaType) \val\true, T) \leftarrow}\\
\quad\happensAt\mathit{(entersArea(Vessel,AreaID),T),}\\
\quad\mathit{areaType(AreaID,AreaType).}\\[2pt]
\terminatedAt\mathit{(withinArea(Vessel,AreaType) \val \true, T) \leftarrow}\\
\quad\happensAt\mathit{(leavesArea(Vessel,AreaID),T),}\\
\quad\mathit{areaType(AreaID,AreaType).}\\
\terminatedAt\mathit{(withinArea(Vessel,AreaType) \val \true, T) \leftarrow}\\
\quad    \happensAt\mathit{(gap\_start(Vessel), T)}.
\end{mysplit}
\end{align} 
Variables start with an upper-case letter, while predicates and constants start with a lower-case letter. 
$\mathit{withinArea(Vessel,AreaType)}$ is a simple fluent indicating that a $\mathit{Vessel}$ is within an area of some type. 
We chose to define $\mathit{withinArea(Vessel,AreaType)}$ as a Boolean fluent, as opposed to a multi-valued one, since areas of different types may overlap.
$\mathit{entersArea(Vessel, AreaID)}$ and $\mathit{leavesArea(Vessel, AreaID)}$ are input events computed at the spatial preprocessing step (see the top part of Table~\ref{tbl:events}), indicating that a $\mathit{Vessel}$ entered (respectively left) an area with $\mathit{AreaID}$.
$\mathit{areaType(AreaID,AreaType)}$ is an atemporal predicate storing the areas of interest of a given dataset.
$\mathit{withinArea(Vessel,AreaType)\val\true}$ is initiated when a $\mathit{Vessel}$ enters an area of $\mathit{AreaType}$, and terminated when the $\mathit{Vessel}$ leaves the area of $\mathit{AreaType}$.
$\mathit{withinArea(Vessel,AreaType)\val\true}$ is also terminated  when the trajectory synopsis generator produces a $\mathit{gap\_start}$ event (see the middle part of Table~\ref{tbl:events}), indicating the beginning of a communication gap (in the subsection that follows we discuss further communication gaps). In this case we chose to make no assumptions about the location of the vessel. 
With the use of rule-set~\eqref{eq:withinArea}, RTEC computes the \emph{maximal intervals} during which a vessel is said to be within an area of some type.

\begin{table}[t]
\caption{Speed-related building blocks.}\label{tbl:speed-related-bb}\vspace{-0.4cm}
\begin{center}
\renewcommand{\arraystretch}{0.9}
\setlength\tabcolsep{2.5pt}
\begin{tabular}{lcc}
\hline\noalign{\smallskip}
\textbf{Fluent} & \textbf{Min Speed}  & \textbf{Max Speed} \\
 & \textbf{(knots)}  & \textbf{(knots)} \\
\noalign{\smallskip}
\hline
\noalign{\smallskip}

$\mathit{stopped(V)}$ & 0 & 0.5 \\[2pt]

$\mathit{lowSpeed(V)}$ & 0.5 & 5 \\[2pt]

$\mathit{movingSpeed(V)\val below}$  & 0.5 & min service speed \\
& & of vessel type \\[2pt]

$\mathit{movingSpeed(V)\val normal}$ & min service speed & max service speed  \\
&  of vessel type &  of vessel type \\[2pt]

$\mathit{movingSpeed(V)\val above}$  & max service speed \\
&  of vessel type & - \\[2pt]

$\mathit{tuggingSpeed(V)}$ & 1.2 & 15 \\[2pt]

$\mathit{trawlingSpeed(V)}$ & 1.0 & 9.0 \\[2pt]
$\mathit{sarSpeed(V)}$      & 2.7 & -   \\[2pt]
\hline
\end{tabular}
\end{center}
\end{table}

\subsubsection{Communication gap}

According to the trajectory synopsis generator, a communication gap takes place when no message has been received from a vessel for at least 30 minutes. 
All numerical thresholds, however, may be tuned (e.g.~by machine learning algorithms) to meet the requirements of the application under consideration.
A communication gap may occur when a vessel sails in an area with no AIS receiving station nearby, or because the transmission power of its transceiver allows broadcasting in a shorter range, or when the transceiver is deliberately turned off. The rules below present a formalisation of communication gap:
\begin{align}
\begin{mysplit}
\label{eq:gap}
\initiatedAt\mathit{(gap(Vessel) \val nearPorts, T)\leftarrow}\\
\quad    \happensAt\mathit{(gap\_start(Vessel), T)},\\
\quad    \holdsAt\mathit{(withinArea(Vessel,nearPorts)=true,T)}.\\
\initiatedAt\mathit{(gap(Vessel) \val farFromPorts,\ T)\leftarrow}\\
\quad    \happensAt\mathit{(gap\_start(Vessel), T)},\\
\quad    \nbf\ \holdsAt\mathit{(withinArea(Vessel,nearPorts)=true,T)}.\\[3pt]
\terminatedAt\mathit{(gap(Vessel) \val \_Value, T)\leftarrow}\\
\quad    \happensAt\mathit{(gap\_end(Vessel), T)}.\\
\end{mysplit}
\end{align} 
$\mathit{gap}$ is a simple, multi-valued fluent, $\mathit{gap\_start}$ and $\mathit{gap\_end}$ are input critical events (see  Table~\ref{tbl:events}), `\nbf' expresses Prolog's negation-by-failure \cite{clark78}, while variables starting with `\_', such as $\mathit{\_Value}$, are free. 
We chose to distinguish between communication gaps occurring near ports from those occurring in the open sea, as the first ones usually do not have a significant role in maritime monitoring. 
According to rule-set \eqref{eq:gap}, a communication gap is said to be initiated when the synopsis generator emits a `gap start' event, and terminated when a `gap end' is detected. 
Given this rule-set, RTEC computes the maximal intervals for which a vessel is not sending position signals.

\subsubsection{Speed-related building blocks}

We have defined a number of speed-related building blocks that are useful in the specification of the more complex maritime patterns, which will be presented in the following section. Table \ref{tbl:speed-related-bb} presents these building blocks.
$\mathit{stopped(V)}$ e.g.~indicates that the vessel $V$ has speed between 0 and 0.5 knots. 
All numerical thresholds have been set in collaboration with domain experts. Moreover, as mentioned earlier, these thresholds may  be optimised for different monitoring applications. $\mathit{stopped(V)}$ is specified by means of the $\mathit{stop\_start}$ and $\mathit{stop\_end}$ input critical events. Similarly, $\mathit{lowSpeed(V)}$ indicates that the vessel $V$ sails at a speed between 0.5 and 5 knots, and is defined by means of the $\mathit{slow\_motion\_start}$ and $\mathit{slow\_motion\_end}$  critical events.
$\mathit{stopped(V)}$ and $\mathit{lowSpeed(V)}$ are independent of the vessel type. In contrast, the multi-valued $\mathit{movingSpeed(V)}$ fluent takes into consideration the minimum and maximum service speed---the speed maintained by a ship under normal load and weather conditions---of the type of vessel $V$. For example, $\mathit{movingSpeed(V)\val normal}$ when $V$ is a cargo vessel and its speed is between 9 and 15 knots.

$\mathit{tuggingSpeed(V)}$ is used for the specification of tugging, i.e.~the activity of pulling a ship into a port,   
$\mathit{trawlingSpeed(V)}$ is restricted to fishing vessels, while  $\mathit{sarSpeed(V)}$ concerns search and rescue (SAR) vessels. For example, $\mathit{sarSpeed(V)}$ is \true\ when the SAR vessel $V$ has speed above 2.7 knots (as indicated in Table \ref{tbl:speed-related-bb}, we have not set an upper speed limit for SAR vessels). 
In addition to the building blocks displayed in Table \ref{tbl:speed-related-bb}, we  employ 
$\mathit{changingSpeed}$, a fluent expressing the maximal intervals during which a vessel is changing its speed, according to the input $\mathit{change\_in\_speed\_start}$ and $\mathit{change\_in\_speed\_end}$ critical events (see Table \ref{tbl:events}).

\begin{figure}[t]
\includegraphics[width=\columnwidth]{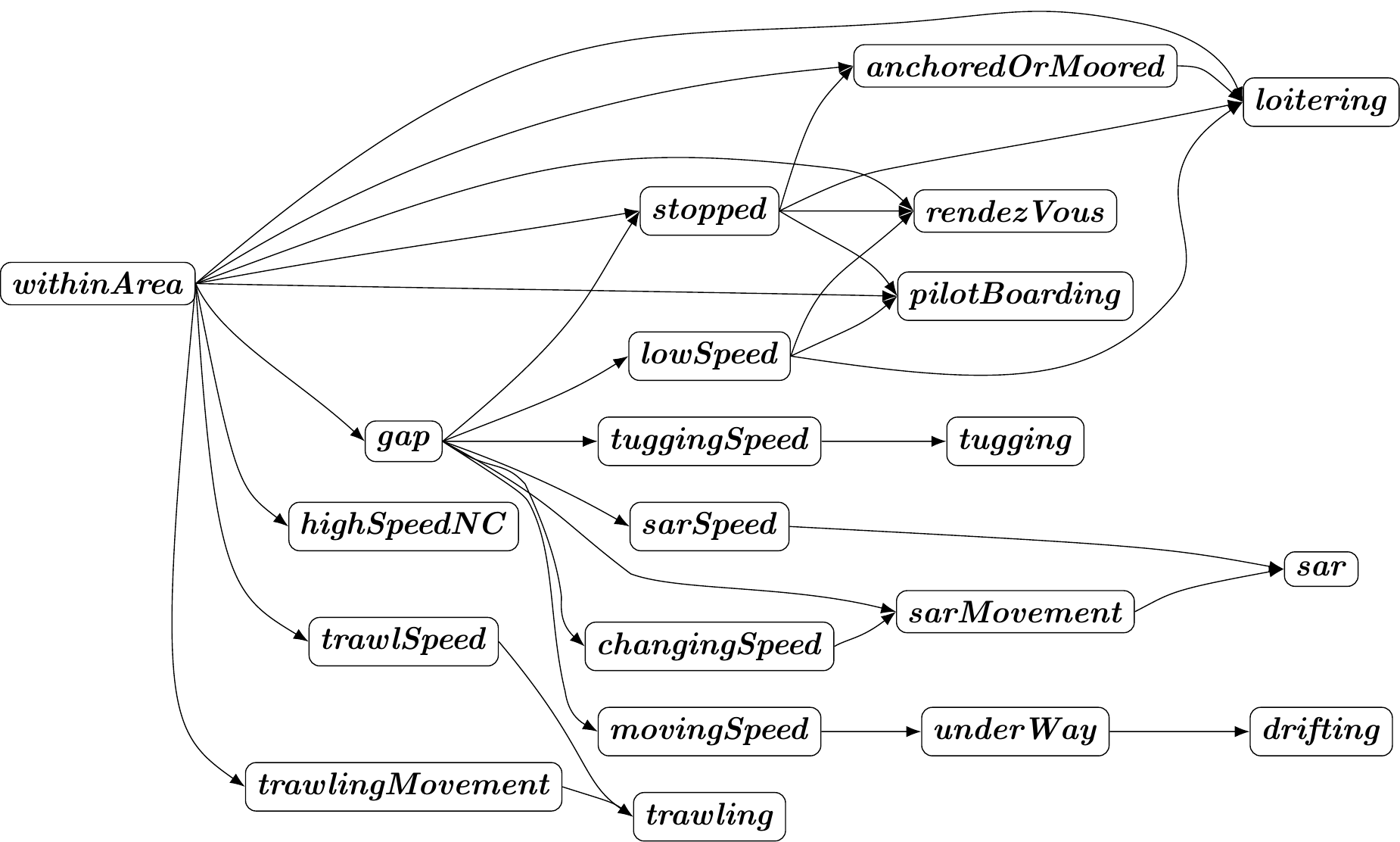}
\caption{Pattern Hierarchy.}
\label{fig:dependency-graph}
\end{figure}

\subsection{Maritime Situational Indicators}
\label{sec:msis}

Our aim is to detect in real-time Maritime Situational Indicators \cite{datacron-del5.1}, i.e.~maritime activities of special significance. The indicators that we have formalised are the synthesis of the outcomes of workshops on user requirements elicitation~\cite{SMARTracIn, 5730278}, restricting attention to AIS data \cite{Ray2019}.
%
%
%
Figure \ref{fig:dependency-graph} displays the hierarchy of our formalisation, i.e.~the relations between the indicators' specifications. In this figure, an arrow from fluent $A$ to fluent $B$ denotes that $A$ is used in the specification of $B$.
To avoid clutter, we omit the presentation of the input stream elements (items above the double horizontal line in Table \ref{tbl:events}) from Figure \ref{fig:dependency-graph}.


\subsubsection{Vessel with high speed near coast}

Several countries have regulated maritime zones. In French territorial waters, for example, there is a 5 knots speed limit for vessels or watercrafts within 300 meters from the coast. One of the causes of marine accidents near the coast is vessels sailing with high speed, thus the early detection of violators ensures safety by improving the efficiency of law enforcement. Figure~\ref{fig:highspeed} illustrates a vessel not conforming to the above regulations. Consider the following formalisation:
\begin{figure}[t]
    \centering
    \includegraphics[width=0.95\columnwidth]{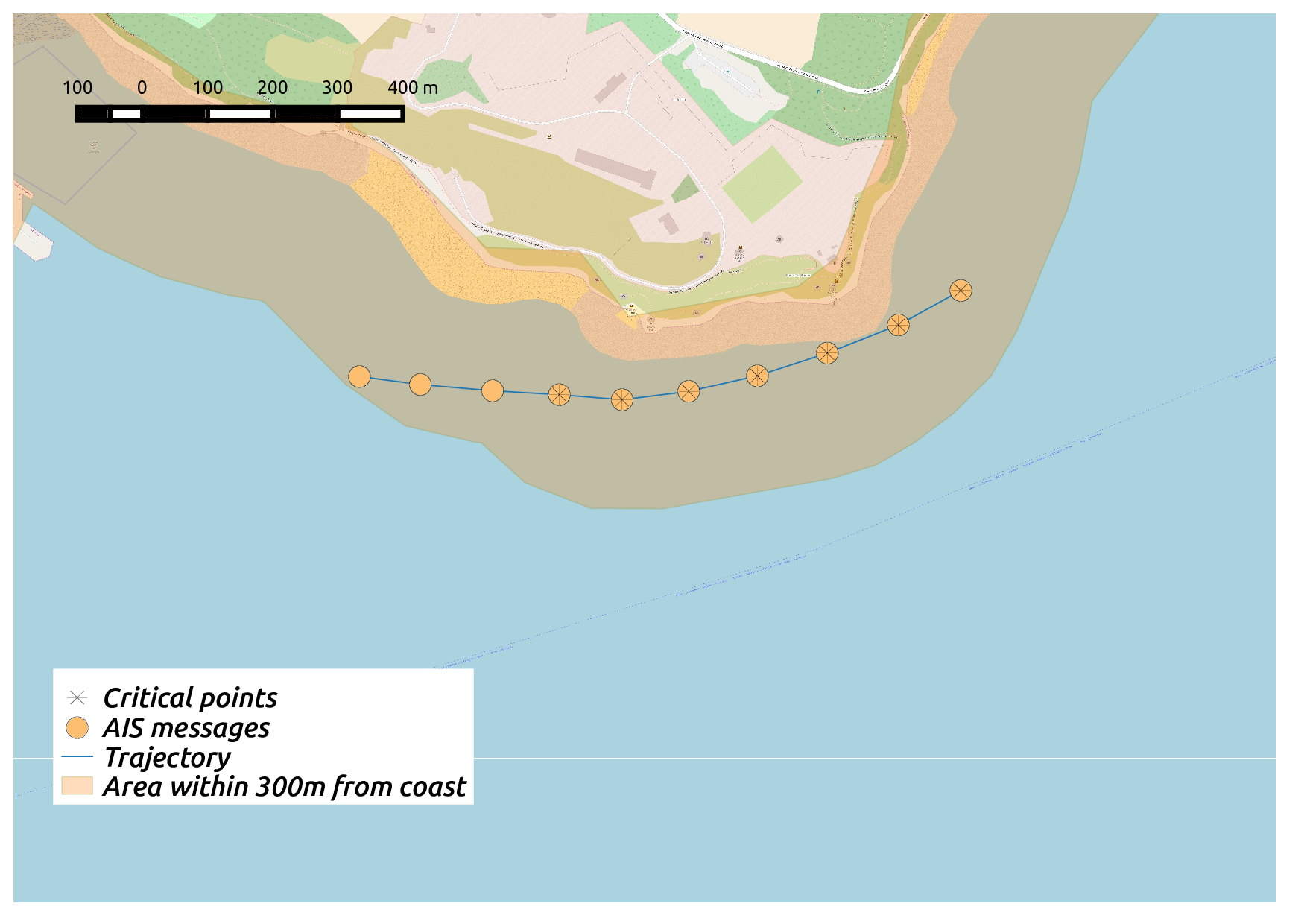}
    \caption{A vessel near the port of Brest, France, with speed above the 5 knots limit. The marked circles denote the AIS position signals that are labelled as `critical' by the synopsis generator. }
    \label{fig:highspeed}
\end{figure}
\begin{align}
\begin{mysplit}
\label{eq:highspeednc}
\initiatedAt\mathit{(highSpeedNC(Vessel) \val \true , T)\leftarrow}\\
\quad    \happensAt\mathit{(velocity(Vessel, Speed, \_CoG, \_TrueHeading),T)},\\
\quad    \holdsAt\mathit{(withinArea(Vessel,nearCoast)\val\true,T),}\\
\quad    \mathit{threshold(v_{hs},V_{hs})},\ \ \ \mathit{Speed>V_{hs}}.\\[3pt]
\terminatedAt\mathit{(highSpeedNC(Vessel) \val \true, T)\leftarrow}\\
\quad    \happensAt\mathit{(velocity(Vessel, Speed, \_CoG, \_TrueHeading),T)},\\
\quad    \mathit{threshold(v_{hs},V_{hs})},\ \ \    \mathit{Speed\leq V_{hs}}.\\
\terminatedAt\mathit{(highSpeedNC(Vessel) \val \true, T)\leftarrow}\\
\quad    \happensAt\mathit{(\endE(withinArea(Vessel,nearCoast)\val\true),T).}
\end{mysplit}
\end{align} 
$\mathit{highSpeedNC(Vessel)}$ is a Boolean simple fluent indicating that a $\mathit{Vessel}$ is exceeding the speed limit imposed near the coast. 
$\mathit{velocity}$ is input contextual information expressing the speed, course over ground (CoG) and true heading of a vessel (the use of the last two parameters will be illustrated in the following sections). This information is attached to each incoming AIS message. 
Recall that variables starting with `\_' are free. 
$\mathit{withinArea(Vessel,nearCoast)\val\true}$ expresses the time periods during which a $\mathit{Vessel}$ is within 300 meters from the French coastline (see rule-set \eqref{eq:withinArea} for the specification of $\mathit{withinArea}$).
$\mathit{threshold}$ is an auxiliary atemporal predicate recording the numerical thresholds of the maritime patterns. The use of this predicate supports code transferability, since the use of different thresholds for different applications requires only the modification of the $\mathit{threshold}$ predicate, and not the modification of the patterns. 
%
%
\endE$(F\val V)$ (respectively~\startE$(F\val V)$) is an RTEC built-in event indicating the ending (resp.~starting) points for which $F\val V$ holds continuously.
According to rule-set \eqref{eq:highspeednc}, therefore, $\mathit{highSpeedNC(Vessel)\val\true}$ is initiated when the $\mathit{Vessel}$ sails within 300 meters from the French coastline with speed above 5 knots, and terminated when its speed goes below 5 knots, sails away (further than 300 meters) from the coastline, or stops sending position signals (recall that $\mathit{withinArea}$ is terminated/\endE ed by $\mathit{gap\_start}$).


\begin{figure}[t]
    \centering
    \includegraphics[width=0.95\columnwidth]{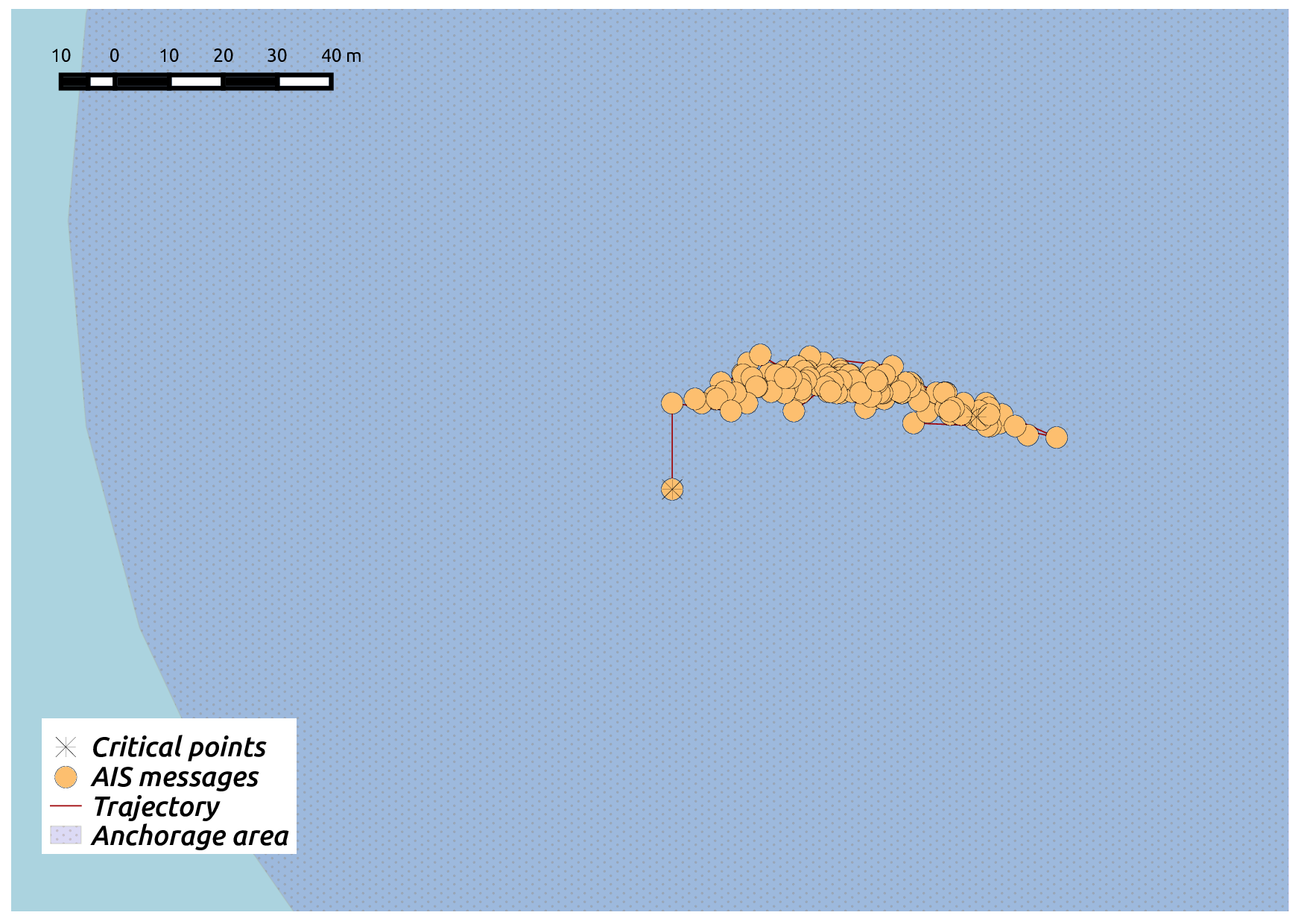}
    \caption{Anchored vessel. 
    }
    \label{fig:anchored}
\end{figure}

\subsubsection{Anchored or moored vessel}

A vessel lowers its anchor in specific areas---e.g.~waiting to enter into a port, or taking on cargo or passengers where insufficient port facilities exist. See Figure~\ref{fig:anchored} for an example of a vessel stopped in an anchorage area. Furthermore, vessels may be moored, i.e.~when a vessel is secured with ropes in any kind of permanent fixture such as a quay or a dock. Consider the specification below:
\begin{align}
\begin{mysplit}
\label{eq:anchoredormoored}
\holdsFor\mathit{(anchoredOrMoored(Vessel)\val\true, I)\leftarrow}\\
\quad    \holdsFor\mathit{(stopped(Vessel)\val farFromPorts,\ I_{sffp})},\\
\quad	 \holdsFor\mathit{(withinArea(Vessel,anchorage)\val true,\ I_{wa})},\\
\quad    \intersectall\mathit{([I_{sffp},I_{wa}],I_{sa})},\\
\quad 	 \holdsFor\mathit{(stopped(Vessel)\val nearPorts,\ I_{sn})},\\
\quad    \unionall\mathit{([I_{sa},I_{sn}],I_i)},\\
\quad    \mathit{threshold(v_{aorm}, V_{aorm}),}\ \ \    \mathit{intDurGreater(I_i, V_{aorm}, I)}.\\
\end{mysplit}
\end{align}
$\mathit{anchoredOrMoored(Vessel)}$ is a statically determined fluent, i.e.~it is specified by means of a domain-dependent \holdsFor\ predicate and interval manipulation constructs---\intersectall\ and \unionall\ in this case, that compute, respectively, the intersection and union of lists of maximal intervals (see Table~\ref{tbl:ec} and Figure \ref{fig:interval-manipulation}). 
Recall that $\mathit{stopped}$ is a fluent recording the intervals in which a vessel is stopped (see Table \ref{tbl:speed-related-bb})---this may be far from all ports or near some port(s). 
$\mathit{intDurGreater(I',V_t,I)}$ is an auxiliary predicate keeping only the intervals $I$ of list $I'$ with length greater than $V_t$.
$\mathit{anchoredOrMoored(Vessel)\val\true}$, therefore, holds when the $\mathit{Vessel}$ is stopped in an anchorage area or near some port, for a time period greater than some threshold (see $\mathit{V_{aorm}}$ in rule \eqref{eq:anchoredormoored}). In our experiments, this threshold was set to 30 minutes. 

\begin{figure}[t]
\centering
\includegraphics[width=0.8\columnwidth]{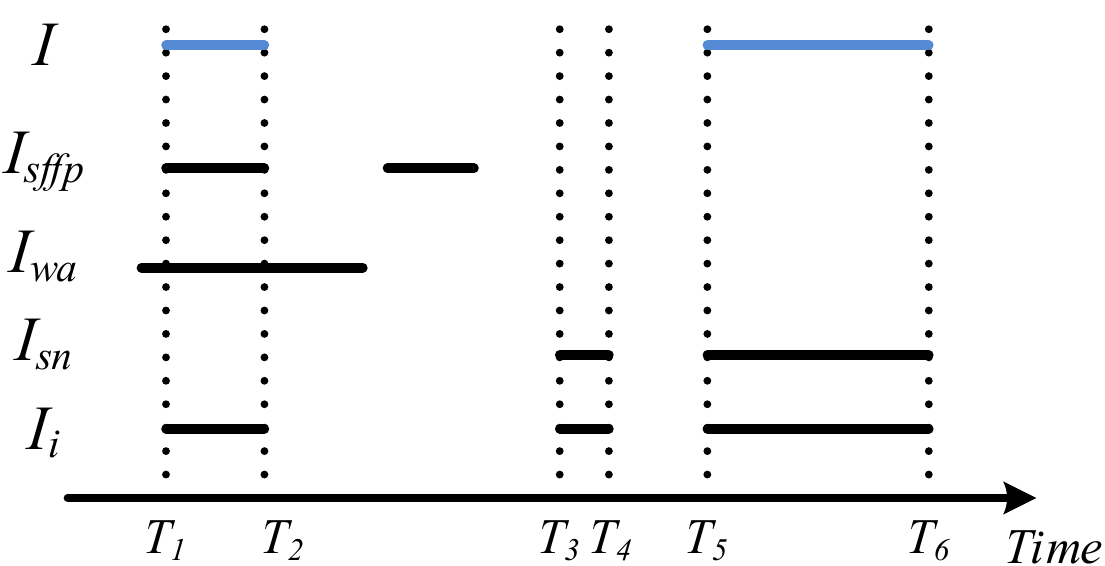}
\caption{Interval computation example of $\mathit{anchoredOrMoored}$. }
\label{fig:anchored-intervals}
\end{figure}

Figure \ref{fig:anchored-intervals} illustrates with the use of a simple example the computation of the $\mathit{anchoredOrMoored}$ intervals. The displayed intervals $I$, $\mathit{I_{sffp}}$, etc, correspond to the intervals of rule \eqref{eq:anchoredormoored}. In the example of Figure \ref{fig:anchored-intervals}, the second interval of $I_i$, $[T_3,T_4]$,  is discarded since it is not long enough according to the $\mathit{V_{aorm}}$ threshold. 

\subsubsection{Drifting vessel}

\begin{figure}[t]
    \centering
    \includegraphics[width=0.95\columnwidth]{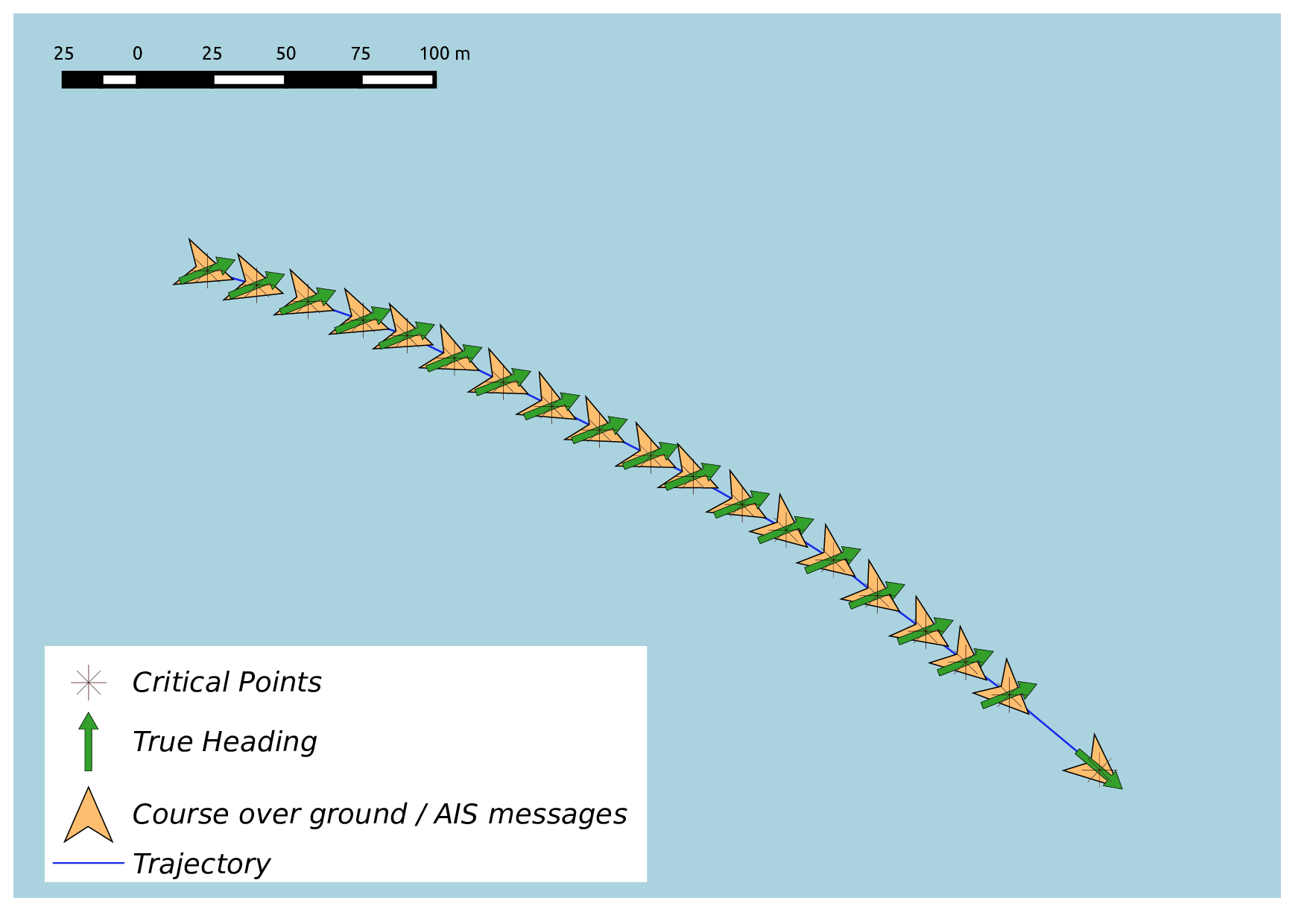}
    \caption{A drifting vessel. In this example, all AIS position signals have been labelled as `critical' ($\mathit{change\_in\_heading}$).}
    \label{fig:drifting}
\end{figure}

A vessel is drifting when its course over ground, i.e.~the direction calculated by the GPS signal, is heavily affected by sea currents or harsh weather conditions. Typically, as illustrated in Figure~\ref{fig:drifting}, when the course over ground deviates from the true heading of a sailing vessel, i.e.~the direction of the ship's bow, then the vessel is considered drifting.
Consider the formalisation below:
\begin{align} 
\begin{mysplit}
\label{eq:drifting}
\initiatedAt\mathit{(drifting(Vessel)\val \true, T)\leftarrow}\\
\quad    \happensAt\mathit{(velocity(Vessel,\_Speed,CoG,TrueHeading),T)},\\
\quad    \mathit{angleDiff(CoG,TrueHeading,Ad)},\\
\quad \mathit{threshold(v_{ad},V_{ad}),\ \ \ Ad>V_{ad}},\\
\quad    \holdsAt\mathit{(underWay(Vessel)\val\true,T)}.\\[3pt]
\terminatedAt\mathit{(drifting(Vessel)\val\true,\ T)\leftarrow}\\
\quad    \happensAt\mathit{(velocity(Vessel,\_Speed,CoG,TrueHeading),T)},\\
\quad    \mathit{angleDiff(CoG,TrueHeading,Ad)},\\
\quad    \mathit{threshold(v_{ad},V_{ad}),\ \ \ Ad\leq V_{ad}}.\\
\terminatedAt\mathit{(drifting(Vessel)\val\true, T)\leftarrow}\\
\quad    \happensAt\mathit{(\endE(underWay(Vessel)\val\true),T)}.
\end{mysplit}
\end{align}
%
$\mathit{drifting}$ is a Boolean simple fluent, while, as mentioned earlier, 
$\mathit{velocity}$ is input contextual information, attached to each AIS message, expressing the speed, course over ground (CoG) and true heading of a vessel.  
$\mathit{angleDiff(A,B,C)}$ is an auxiliary predicate calculating the absolute minimum difference $C$ between two angles $A$ and $B$.
The intervals during which $\mathit{underWay(V)}\val\true$ are computed by the union of the intervals during which the vessel is moving, i.e.~$\mathit{movingSpeed(V)\val below}$, $\mathit{movingSpeed(V)\val normal}$ and $\mathit{movingSpeed(V)\val above}$ (see Table \ref{tbl:speed-related-bb}).
The use of $\mathit{underWay}$ in the initiation and termination conditions of $\mathit{drifting}$ (see rule-set \eqref{eq:drifting}) expresses the constraint that only moving vessels can be considered to be drifting.

\begin{figure}[t]
    \centering
    \includegraphics[width=0.95\columnwidth]{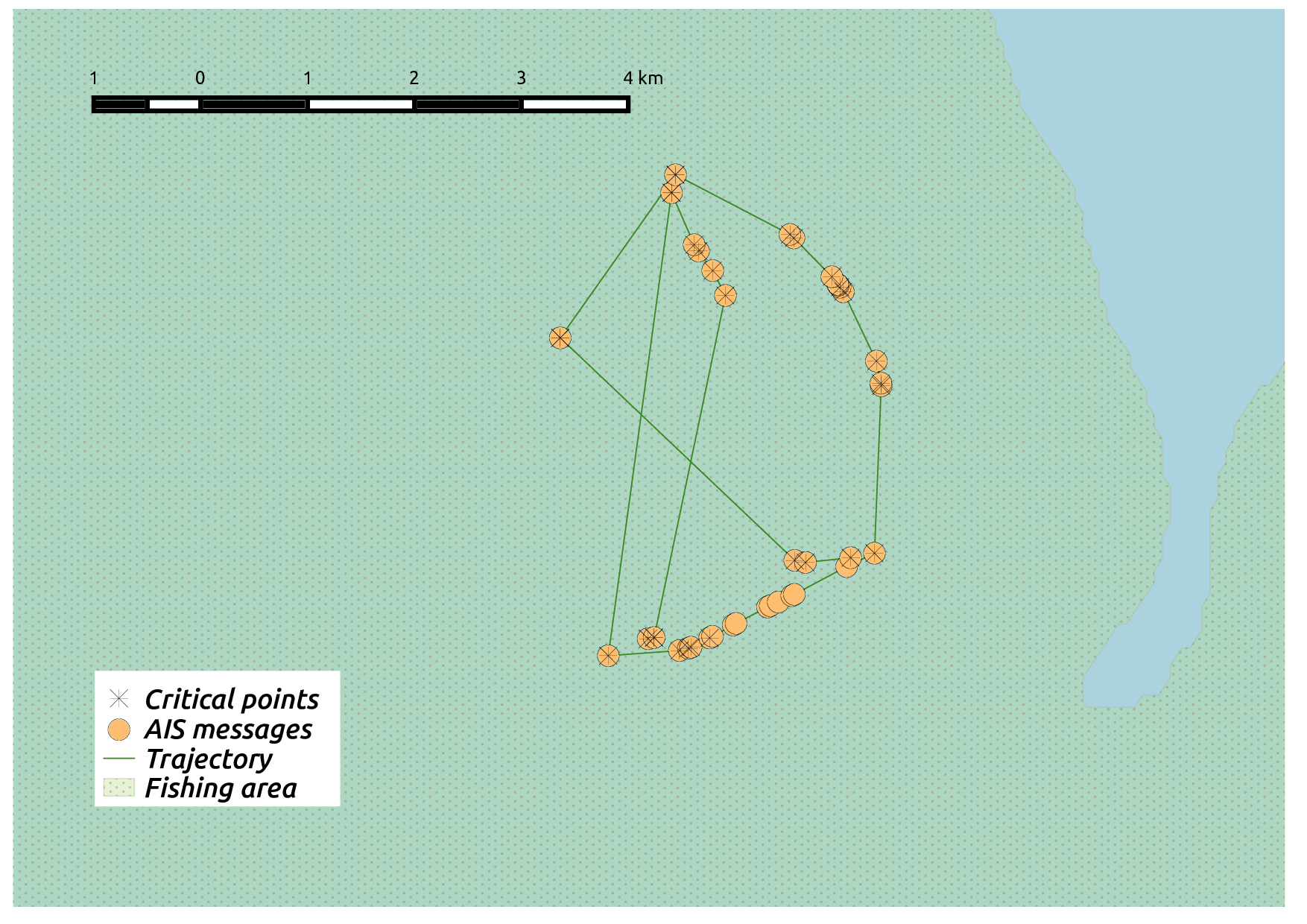}
    \caption{A fishing vessel engaged in trawling. }
    \label{fig:trawlingimg}
\end{figure}

\subsubsection{Trawling}

Fishing is an activity that exploits natural resources, and thus needs to be regulated to safeguard fair access and sustainability~\cite{datacron-del5.1}. Maritime monitoring enables better regulation and monitoring of fishing activities.
A common fishing method is trawling, involving a boat---trawler---pulling a fishing net through the water behind it. The trawler has steady---trawling---speed and a wide heading angle distribution. See Figure~\ref{fig:trawlingimg} for an illustration.
A formalisation of trawling movement may be found below:
\begin{align} 
\begin{mysplit}
\label{eq:trawlingcrs}
\initiatedAt\mathit{(trawlingMovement(Vessel)\val\true, T)\leftarrow}\\
\quad    \happensAt\mathit{(change\_in\_heading(Vessel),T)},\\
\quad    \mathit{vesselType(Vessel,fishing),}\\
\quad    \holdsAt\mathit{(withinArea(Vessel,fishing)\val \true),T)}.\\[2pt]
\terminatedAt\mathit{(trawlingMovement(Vessel)\val\true, T)\leftarrow}\\
\quad    \happensAt\mathit{(\endE(withinArea(Vessel,fishing)\val \true),T)}.
\end{mysplit}
\end{align}
$\mathit{trawlingMovement}$ is defined as a simple fluent, $\mathit{change\_in\_heading}$ is an input critical event detected at the trajectory synopsis step, and $\mathit{vesselType}$ is an auxiliary atemporal predicate recording the vessel types of a given dataset. 
$\mathit{trawlingMovement}$ is subject to the `deadlines' mechanism of RTEC, i.e.~this fluent is automatically terminated after a designated period of time---10 minutes in the experiments presented in the following section---has elapsed since its last initiation.
(We omit the corresponding RTEC declarations to simplify the presentation.)
As shown in rule-set \eqref{eq:trawlingcrs}, $\mathit{trawlingMovement}$ is also terminated when the vessel in question leaves the fishing area.
(In other applications, it may be desirable to relax the constraint of restricting attention to designated fishing areas.)
Consequently, $\mathit{trawlingMovement(Vessel)}$ is \true\ as long as  the fishing $\mathit{Vessel}$ performs a sequence of heading changes, each taking place at the latest 10 minutes after the previous one, while sailing in a fishing area. 
Trawling can then be specified  as follows:
\begin{align} 
\begin{mysplit}
\label{eq:trawling}
\holdsFor\mathit{(trawling(Vessel) \val \true, I)\leftarrow}\\
\quad    \holdsFor\mathit{(trawlingMovement(Vessel) \val \true, I_{tc}}),\\
\quad    \holdsFor\mathit{(trawlSpeed(Vessel) \val \true, I_t)},\\
\quad    \intersectall\mathit{([I_{tc},I_t], I_i)},\\
\quad    \mathit{threshold(v_{trawl},V_{trawl}),}\ \ \    \mathit{intDurGreater(I_i,V_{trawl},I)}.\\
\end{mysplit}
\end{align}
$\mathit{trawling}$ is a statically determined fluent and
$\mathit{trawlSpeed}$ is a simple fluent recording the intervals during which a vessel sails at trawling speed (see Table \ref{tbl:speed-related-bb}).
According to rule \eqref{eq:trawling}, therefore, a vessel is said to be trawling if it is a fishing vessel, has trawling movement and sails in trawling speed for a period of time greater than $\mathit{V_{trawl}}$ (1 hour in our experiments).


\subsubsection{Tugging}

\begin{figure}[t]
    \centering
    \includegraphics[width=0.95\columnwidth]{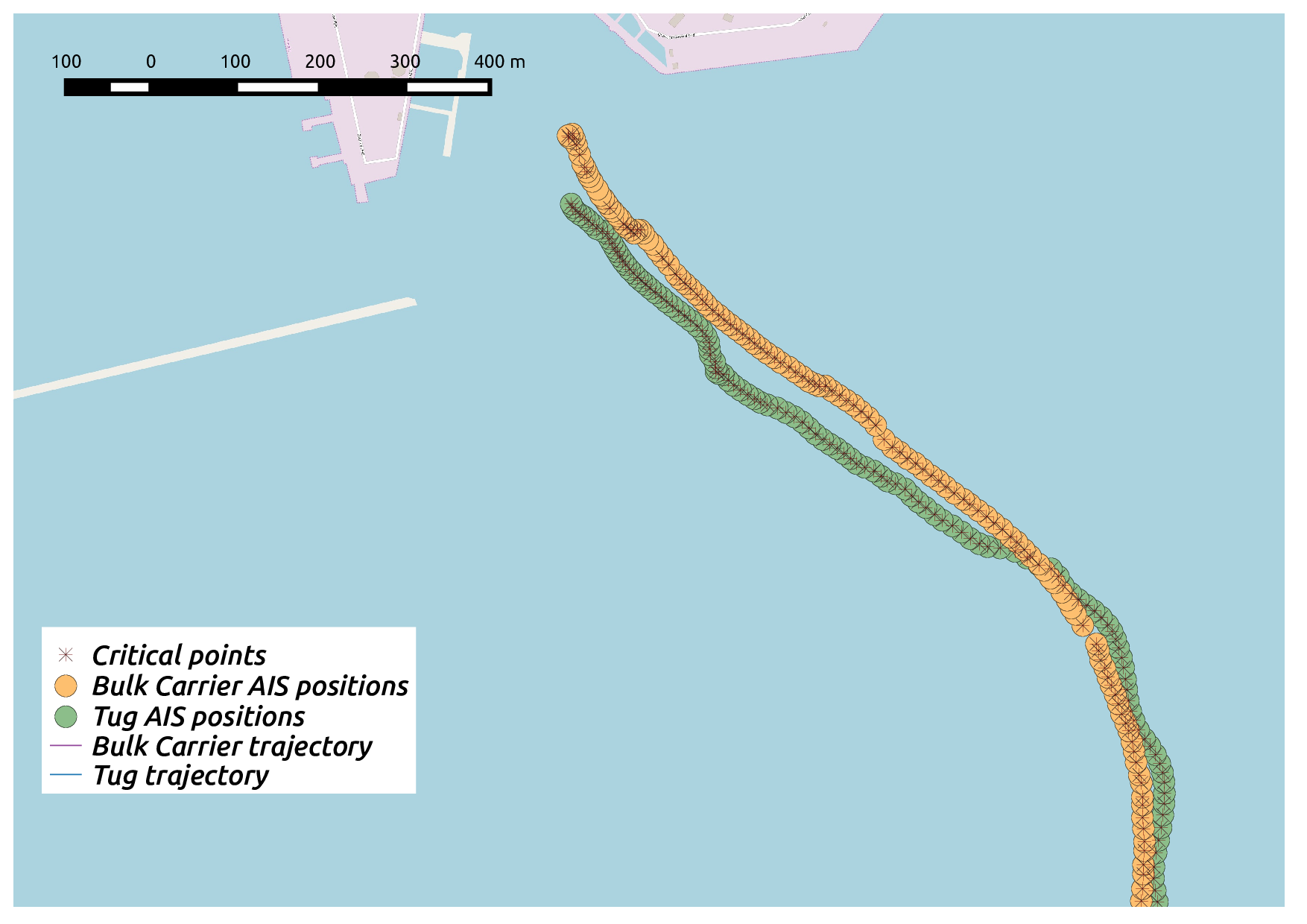}
    \caption{Example of bulk carrier tugging. In this case, all position signals are labelled as `critical'. }
    \label{fig:tuggedimg}
\end{figure}

A vessel that should not move by itself---e.g.~a ship in a crowded harbor or a narrow canal---or a vessel that cannot move by itself is typically pulled or towed by a tug boat. 
Figure~\ref{fig:tuggedimg} shows an example. It is expected that during tugging the two vessels are close and their speed is lower than normal, for safety and manoeuvrability reasons.
We have formalised tugging as follows:
\begin{align} 
\begin{mysplit}
\label{eq:tugging}
\holdsFor\mathit{(tugging(Vessel_1, Vessel_2) \val \true, I) \leftarrow}\\
\quad    \mathit{oneIsTug(Vessel_1,Vessel_2)},\ \ \ \nbf\ \mathit{oneIsPilot(Vessel_1,Vessel_2)},\\
\quad    \holdsFor\mathit{(proximity(Vessel_1, Vessel_2) \val \true,\ I_p)},\\
\quad    \holdsFor\mathit{(tuggingSpeed(Vessel_1)\val\true,I_{ts1}),}\\
\quad    \holdsFor\mathit{(tuggingSpeed(Vessel_2)\val\true,I_{ts2}),}\\
\quad	\intersectall\mathit{([I_p,I_{ts1},I_{ts2}],I_i),}\\
\quad    \mathit{threshold(v_{tug},V_{Tug}),}\ \ \    \mathit{intDurGreater(I_i,V_{Tug},I)}.\\
\end{mysplit}
\end{align}
$\mathit{tugging}$ is a relational fluent referring to a pair of vessels, as opposed to the fluents presented so far that concern a single vessel. 
$\mathit{oneIsTug(V_1,V_2)}$ is an auxiliary predicate stating whether one of vessels $V_1, V_2$ is a tug boat.
Similarly, $\mathit{oneIsPilot(V_1, V_2)}$ states whether one of vessels $V_1, V_2$ is a pilot boat. Piloting will be discussed shortly.
$\mathit{proximity}$ is a durative input fluent computed at the spatial preprocessing step (see Table \ref{tbl:events}), expressing the time periods during which two vessels are `close' (in the presented experiments, their distance is less than 100 meters).
$\mathit{tuggingSpeed}$ is a simple fluent expressing the intervals during which a vessel is said to be sailing at tugging speed (see Table \ref{tbl:speed-related-bb}).
According to rule \eqref{eq:tugging}, two vessels are said to be engaged in tugging if one of them is a tug boat, neither of them is a pilot boat, and, for at least $\mathit{V_{Tug}}$ time-points, they are close to each other and sail at tugging speed.


\subsubsection{Piloting}

During piloting, a highly experienced sailor in navigation in specific areas---a maritime pilot---approaches with a pilot boat, boards and manoeuvres another vessel through dangerous or congested areas. 
Maritime pilots are navigational experts with knowledge of a particular area such as its depth, currents and hazards.
Piloting, therefore, is of major importance for maritime safety. 
A formalisation of pilot boarding may be found below:
\begin{align} 
\begin{mysplit}
\label{eq:pilotBoarding}
\holdsFor\mathit{(pilotBoarding(Vessel_1, Vessel_2) \val \true, I) \leftarrow}\\
\quad    \mathit{oneIsPilot(Vessel_1,Vessel_2)},\ \ \ \nbf\ \mathit{oneIsTug(Vessel_1,Vessel_2)},\\
\quad    \holdsFor\mathit{(proximity(Vessel_1, Vessel_2) \val \true, I_p)},\\
\quad    \holdsFor\mathit{(lowSpeed(Vessel_1) \val true, I_{l1})},\\
\quad    \holdsFor\mathit{(stopped(Vessel_1) \val farFromPorts, I_{s1})},\\
\quad    \unionall\mathit{([I_{l1}, I_{s1}], I_1)},\\
\quad    \holdsFor\mathit{(lowSpeed(Vessel_2) \val true, I_{l2})},\\
\quad    \holdsFor\mathit{(stopped(Vessel_2) \val farFromPorts, I_{s2})},\\
\quad    \unionall\mathit{([I_{l2}, I_{s2}], I_2)},\\
\quad    \intersectall\mathit{([I_1, I_2, I_p], I_f)},\\
\quad    \holdsFor\mathit{(withinArea(Vessel_1,nearCoast) \val true, I_{nc1})},\\
\quad    \holdsFor\mathit{(withinArea(Vessel_2,nearCoast) \val true, I_{nc2})},\\
\quad    \complementall\mathit{(I_f,[I_{nc1},I_{nc2}],I_i)},\\
\quad    \mathit{threshold(v_{pil}, V_{pil}),}\ \ \    \mathit{intDurGreater(I_i,V_{pil},I)}.\\
\end{mysplit}
\end{align}
$\mathit{pilotBoarding}$ is a relational fluent referring to a pair of vessels, while 
$\mathit{lowSpeed}$ is a  fluent recording the intervals during which a vessel sails at low speed (see Table \ref{tbl:speed-related-bb}).  
\complementall\ is an interval manipulation construct of RTEC (see Table \ref{tbl:ec} and Figure \ref{fig:interval-manipulation}).
According to rule \eqref{eq:pilotBoarding}, $\mathit{pilotBoarding(V_1,V_2)}$ holds when one of the two vessels $V_1, V_2$ is a pilot vessel, neither of them is a tug boat, $V_1, V_2$ are close to each other, and they are stopped or sail at low speed far from the coast. According to domain experts, the boarding procedure in pilot operations takes place far from the coast for safety reasons.

\begin{figure}[t]
    \centering
    \includegraphics[width=0.95\columnwidth]{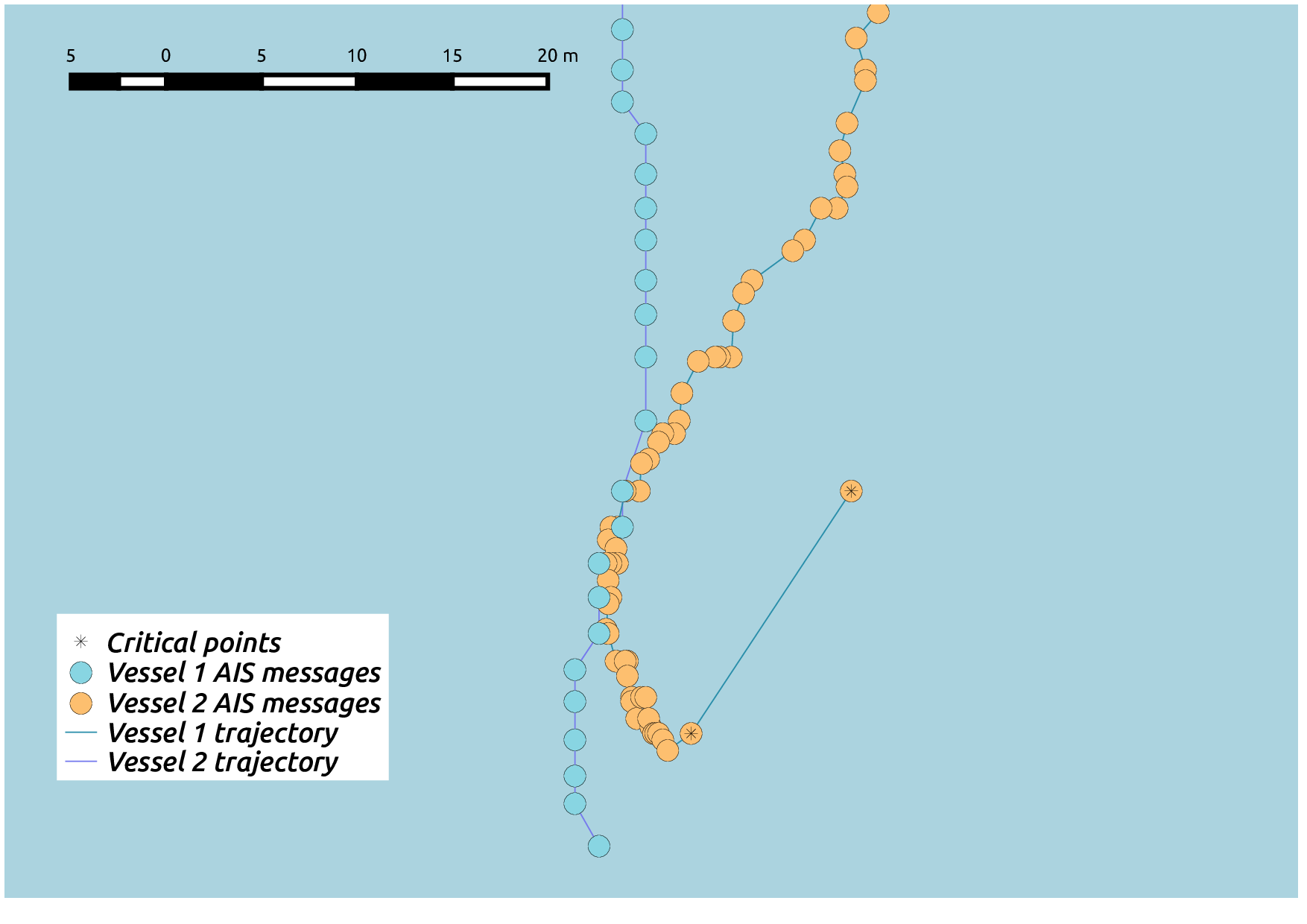}
    \caption{Fishing vessels in close proximity. In this example, the vessels started sailing at a low speed before they came close to each other. Hence, these critical events ($\mathit{slow\_motion\_start}$) are not displayed in the figure.}
    \label{fig:rendezvous}
\end{figure}

\subsubsection{Vessel rendez-vous}

A scenario that may indicate illegal activities, such as illegal cargo transfer, is when two vessels are nearby in the open sea, stopped or sailing at a low speed. See Figure~\ref{fig:rendezvous} for an illustration. A specification of `rendez-vous', or `ship-to-ship transfer', may be found below:
\begin{align} 
\begin{mysplit}
\label{eq:rendezVous}
\holdsFor\mathit{(rendezVous(Vessel_1, Vessel_2) \val \true, I) \leftarrow}\\
\quad    \mathit{\nbf\ oneIsTug(Vessel_1,Vessel_2)},\ \ \    \mathit{\nbf\ oneIsPilot(Vessel_1,Vessel_2)},\\
\quad    \holdsFor\mathit{(proximity(Vessel_1, Vessel_2) \val \true, I_p)},\\
\quad    \holdsFor\mathit{(lowSpeed(Vessel_1) \val true, I_{l1})},\\
\quad    \holdsFor\mathit{(stopped(Vessel_1) \val farFromPorts, I_{s1})},\\
\quad    \unionall\mathit{([I_{l1}, I_{s1}], I_1)},\\
\quad    \holdsFor\mathit{(lowSpeed(Vessel_2) \val true, I_{l2})},\\
\quad    \holdsFor\mathit{(stopped(Vessel_2) \val farFromPorts, I_{s2})},\\
\quad    \unionall\mathit{([I_{l2}, I_{s2}], I_2)},\\
\quad    \intersectall\mathit{([I_1, I_2, I_p], I_f)},\\
\quad    \holdsFor\mathit{(withinArea(Vessel_1,nearPorts) \val true, I_{np1})},\\
\quad    \holdsFor\mathit{(withinArea(Vessel_2,nearPorts) \val true, I_{np2})},\\
\quad    \holdsFor\mathit{(withinArea(Vessel_2,nearCoast) \val true, I_{nc1})},\\
\quad    \holdsFor\mathit{(withinArea(Vessel_2,nearCoast) \val true, I_{nc2})},\\
\quad    \complementall\mathit{(I_f,[I_{np1},I_{np2},I_{nc1},I_{nc2}],I_i)},\\
\quad    \mathit{threshold(v_{rv}, V_{rv}),}\ \ \   \mathit{intDurGreater(I_i,V_{rv},I)}.\\
\end{mysplit}
\end{align}
The above formalisation is similar to that of pilot boarding. 
The differences are the following. 
First, the specification of $\mathit{rendezVous}$ excludes pilot vessels (and tug boats). Second, we require that both vessels are far from ports, as two slow moving or stopped vessels near some port would probably be moored or departing from the port. 
Similar to pilot boarding, we require that the two vessels are not near the coastline. The rationale in this case is that illegal ship-to-ship transfer typically takes place far from the coast. 
Note that, depending on the chosen distance thresholds for $\mathit{nearCoast}$ and $\mathit{nearPorts}$, a vessel may be `far' from the coastline and at the same time `near' some port. 
Moreover, a vessel may be `far' from all ports and `near' the coastline.

\begin{figure}[t]
    \centering
    \includegraphics[width=0.95\columnwidth]{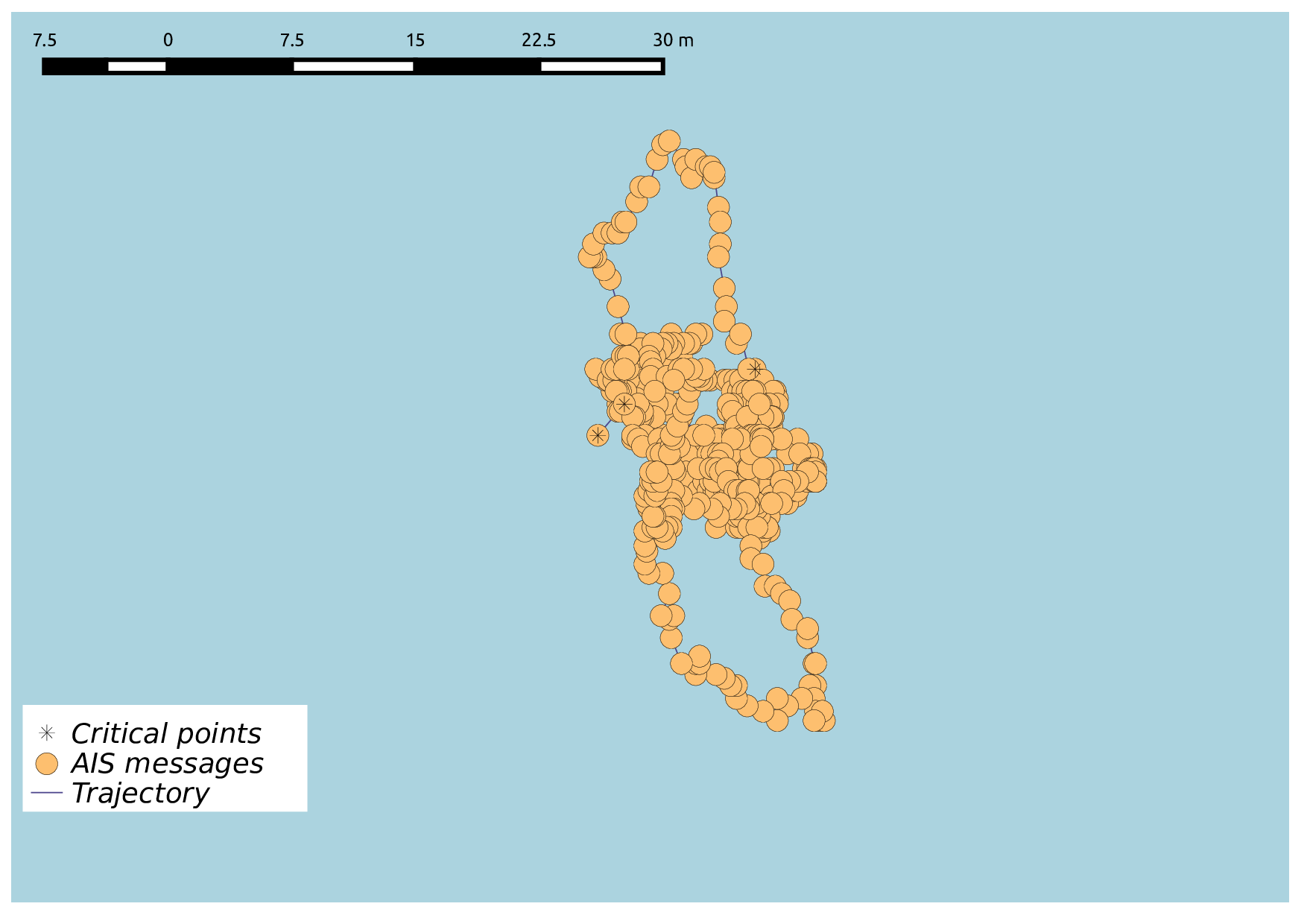}
    \caption{A vessel loitering.}
    \label{fig:loitering}
\end{figure}

\subsubsection{Loitering}
 
Loitering is the act of remaining in a particular area for a long period without any evident purpose. Figure~\ref{fig:loitering} presents an illustration. In sea, this behaviour is an indicator of a potentially unlawful activity. Consider the formalisation below:
\begin{align} 
\begin{mysplit}
\label{eq:loitering}
\holdsFor\mathit{(loitering(Vessel) \val \true, I) \leftarrow}\\
\quad    \holdsFor\mathit{(stopped(Vessel) \val farFromPorts, I_{s})},\\
\quad    \holdsFor\mathit{(lowSpeed(Vessel) \val \true, I_l)},\\
\quad    \unionall\mathit{([I_s,I_l],I_{ls})},\\
\quad    \holdsFor\mathit{(withinArea(Vessel,nearCoast) \val \true, I_{wa})},\\
\quad    \holdsFor\mathit{(anchoredOrMoored(Vessel) \val true, I_{am})},\\
\quad    \complementall\mathit{(I_{ls},[I_{am},I_{wa}],I_{i})},\\
\quad    \mathit{threshold(v_{ltr}, V_{ltr})},\ \ \    \mathit{intDurGreater(I_i, V_{ltr}, I)}.\\
\end{mysplit}
\end{align}
According to rule~\eqref{eq:loitering}, a vessel is said to loiter when it is stopped or sails at a low speed far from ports and the coastline, and it is not anchored, for a period greater than $\mathit{V_{ltr}}$ time-points. For the experiments presented in Section~\ref{sec:analysis}, we have set $\mathit{V_{ltr}}$ to 30 minutes.
%


\subsubsection{Search and rescue operations}

\begin{figure}[t]
    \centering
    \includegraphics[width=0.95\columnwidth]{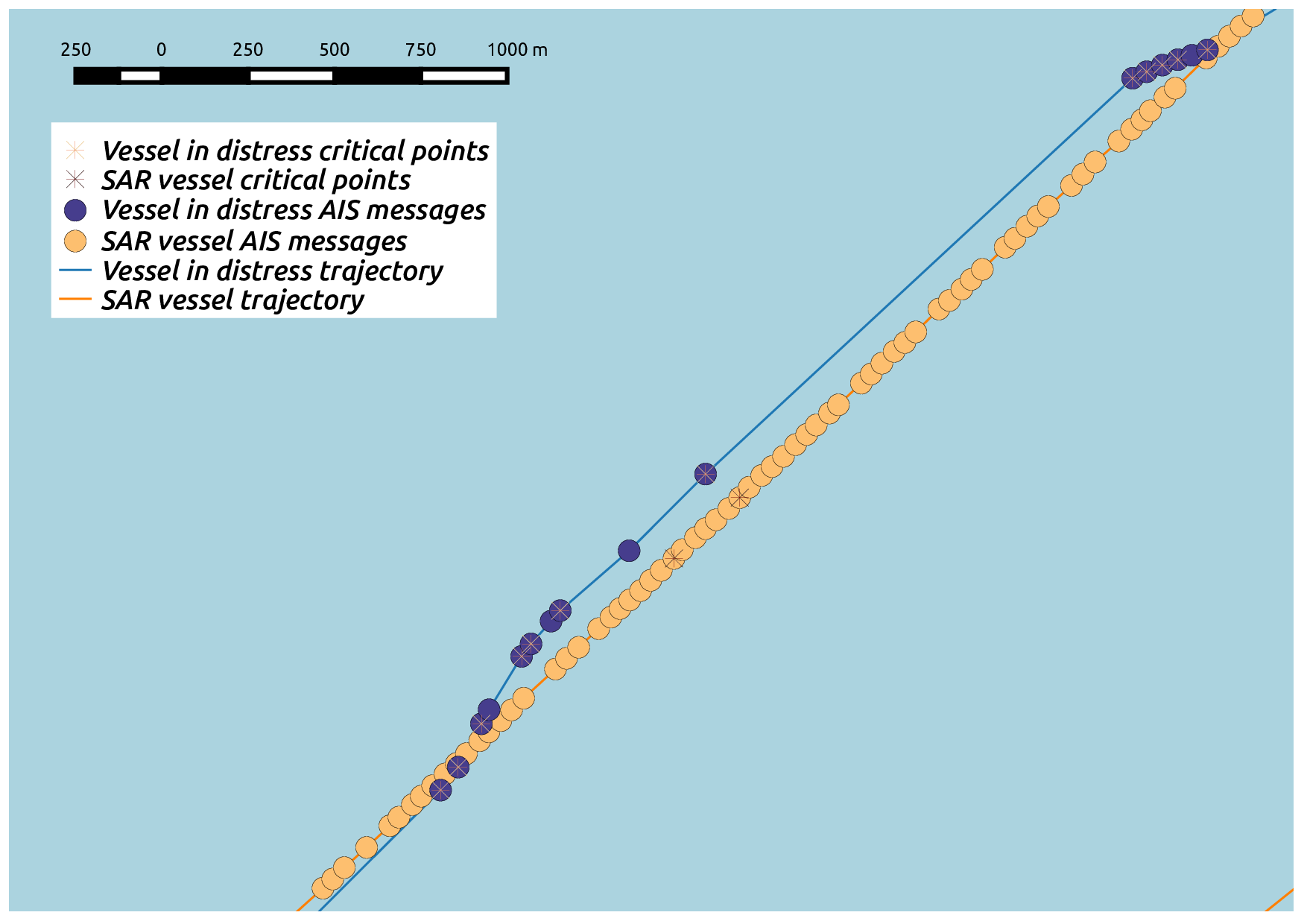}
    \caption{A vessel in distress and a SAR vessel.
    }
    \label{fig:sarimg}
\end{figure}

Search and rescue (SAR) operations aim to provide aid to people who are in distress or imminent danger. See Figure~\ref{fig:sarimg} for an illustration of such an operation. Research has indicated that vessels engaged in SAR operations change speed and heading more often compared to other voyages \cite{chatzikokolakis_konstantinos_2018_1203886}. Taking this into consideration, we developed the formalisation below:
\begin{align} 
\begin{mysplit}
\label{eq:sarcrs}
\initiatedAt\mathit{(sarMovement(Vessel)\val\true,\ T)\leftarrow}\\
\quad\happensAt\mathit{(change\_in\_heading(Vessel),T)},\\
\quad\mathit{vesselType(Vessel,sar).}\\
\initiatedAt\mathit{(sarMovement(Vessel)\val\true,\ T)\leftarrow}\\
\quad\happensAt\mathit{(\startE(changingSpeed(Vessel)\val\true),T)},\\
\quad\mathit{vesselType(Vessel,sar).}\\
\end{mysplit}
\end{align}
$\mathit{sarMovement}$ is an auxiliary fluent recording the intervals of potential SAR operations.
$\mathit{change\_in\_heading}$ is an input critical event, while
$\mathit{changingSpeed}$ is a fluent expressing the maximal intervals during which a vessel is changing its speed, according to the input $\mathit{change\_in\_speed\_start}$ and $\mathit{change\_in\_speed\_end}$ critical events (see Table \ref{tbl:events}). \startE$(F\val V)$ is a built-in event of RTEC indicating the starting points of the maximal intervals during which $F\val V$ holds continuously. 
According to rule-set~\eqref{eq:sarcrs}, $\mathit{sarMovement}$ is initiated when a SAR vessel changes heading or starts changing speed. 
Similar to $\mathit{trawlingMovement}$ (see rule~\eqref{eq:trawlingcrs}), $\mathit{sarMovement}$ is subject to the `deadlines' mechanism of RTEC, i.e.~it is automatically terminated at the latest at some---30 minutes, in our experiments---time-points after its last initiation.
We use the auxiliary $\mathit{sarMovement}$ fluent to express SAR operations as follows:
\begin{align} 
\begin{mysplit}
\label{eq:sar}
\holdsFor\mathit{(sar(Vessel)\val\true, I)\leftarrow}\\
\quad    \holdsFor\mathit{(sarSpeed(Vessel)\val\true, I_s)},\\
\quad    \holdsFor\mathit{(sarMovement(Vessel)\val\true, I_m)},\\
\quad    \intersectall\mathit{([I_s,I_m],I_i) },\\
\quad    \mathit{threshold(v_{sar},V_{sar}),}\ \ \ \mathit{intDurGreater(I_i,V_{sar},I)}.\\
\end{mysplit}
\end{align}
$\mathit{sar(Vessel)}$ is a statically determined fluent denoting that $\mathit{Vessel}$ is engaged in a SAR operation. $\mathit{sarSpeed}$ expresses the speed range of SAR vessels (see Table \ref{tbl:speed-related-bb}). According to rule~\eqref{eq:sar}, $\mathit{sar(Vessel)}$ is \true\ when the $\mathit{Vessel}$ has the speed and movement of a vessel in a SAR operation for at least $\mathit{V_{sar}}$ time-points (1 hour in our experiments).
In the future, we aim to develop a relational formalisation of SAR operations, including a representation of the vessel in distress.

\section{Experimental Analysis}\label{sec:analysis}

We present an evaluation of our maritime monitoring system in terms of efficiency and accuracy using the real-world datasets of the datAcron project. 

\subsection{Experimental Setup}

\begin{figure}[t]
\includegraphics[width=0.9\columnwidth]{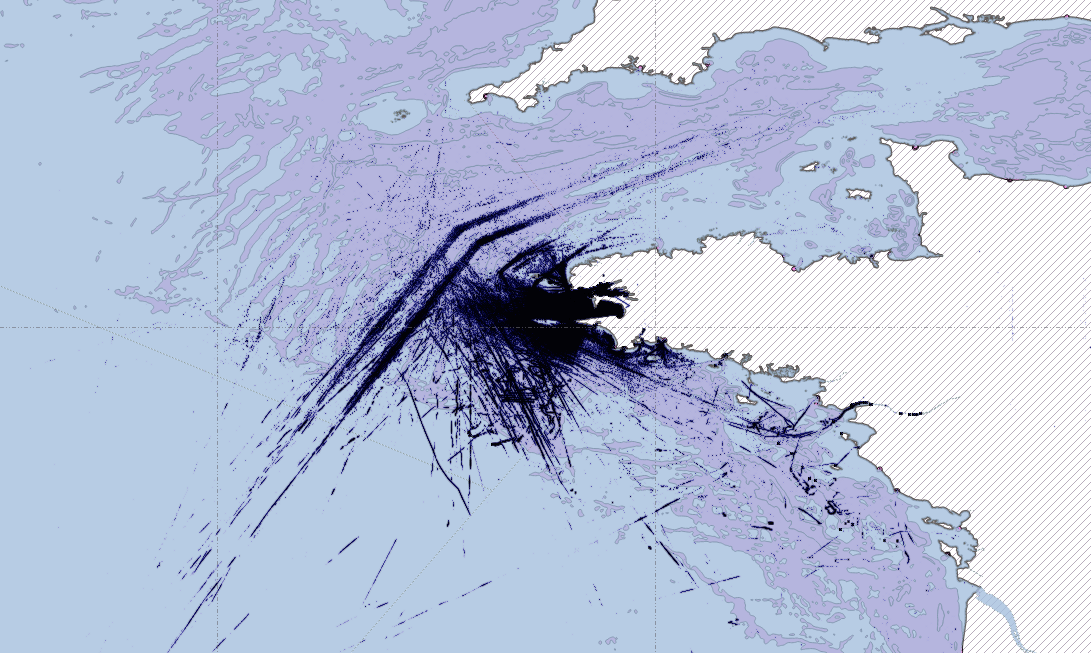}\\~\\
\includegraphics[width=\columnwidth]{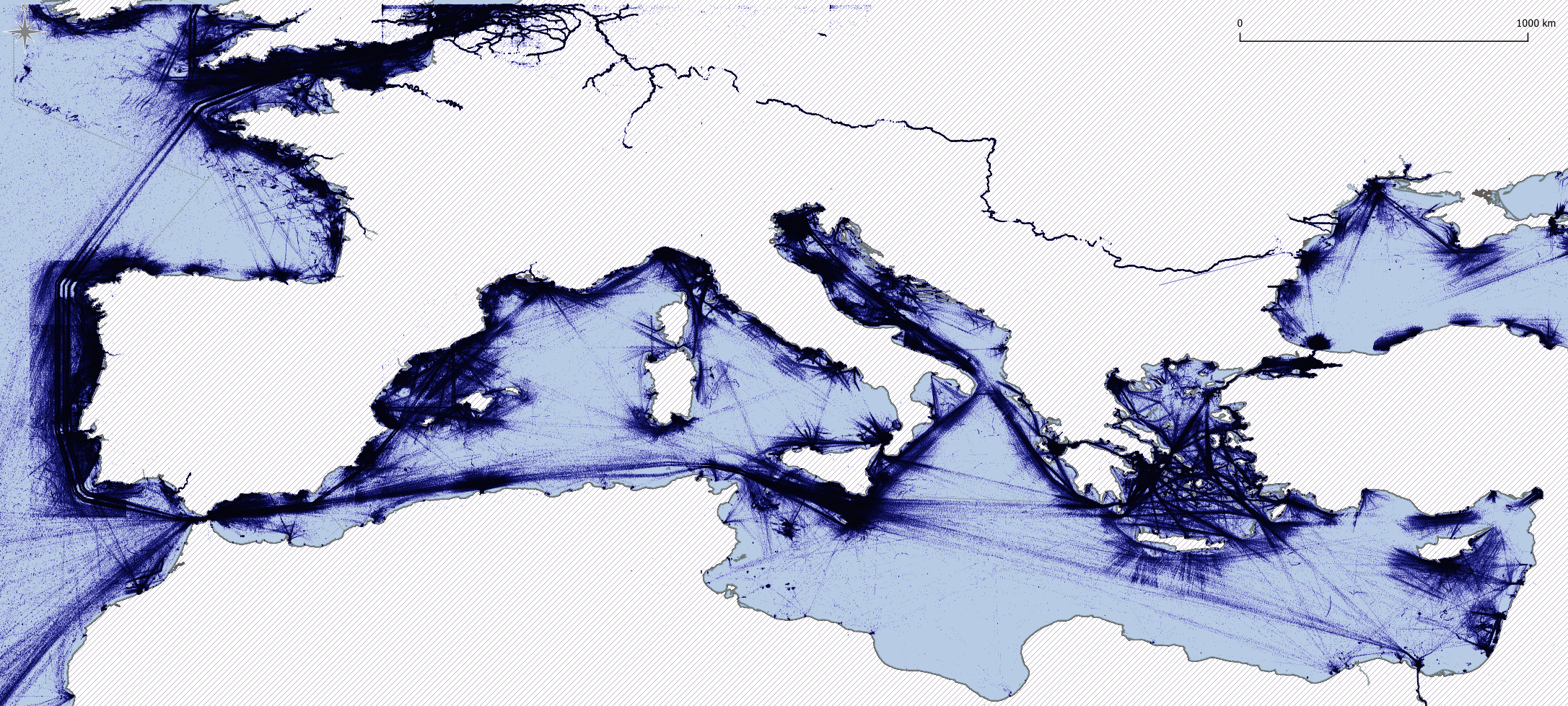}
\caption{Vessel positions from surveillance (AIS) streams---top: Brest, bottom: Europe.}
\label{fig:bresteucov}
\end{figure}

Figure~\ref{fig:bresteucov} illustrates the geographical coverage of the datasets, while Table \ref{tbl:datasets} outlines their characteristics. A summary of the datasets is presented below.

\begin{table}[t]
\centering
\caption{Datasets' characteristics.}
\label{tbl:datasets}\vspace{-0.4cm}
\renewcommand{\arraystretch}{0.9}
\setlength\tabcolsep{4pt}
\begin{tabular}{@{}lcc@{}}
\hline\noalign{\smallskip}
\textbf{Attribute}                    				& \textbf{Brest} & \textbf{Europe}        \\ 
\noalign{\smallskip}
\hline
\noalign{\smallskip}
Period (months)       	  & 6     & 1    \\ [2pt]
Vessels                   & 5K    & 34K  \\[2pt]
Position signals          & 18M   & 55M  \\[2pt]
Spatio-temporal events    & 374K  & 11M   \\[2pt]
Critical position signals & 4.6M  & 17M  \\[2pt]
Fishing areas             & 263   & 1K   \\[2pt]
Natura 2000 areas         & 1.2K  & 2.6K   \\[2pt]
Anchorage areas           & 9     & 9   \\[2pt]
Ports                     & 222   & 2201 \\
\hline
\end{tabular}
\end{table}


\textbf{Brest, France. }
We employed a publicly available dataset consisting of 18M  position signals from 5K vessels sailing in the Atlantic Ocean around the port of Brest, France, between October 1st 2015 and 31st March 2016~\cite{RayData}. The spatial preprocessing module produced 374K spatio-temporal events linking vessels with, among others, 263 fishing areas and 1.2K Natura 2000 areas \cite{doi:10.1080/17445647.2016.1195299,10.1371/journal.pone.0130746}, as well as vessels with other vessels. The trajectory synopsis generator labelled 4.6M position signals as critical. 

\textbf{Europe. }
IMIS Global\footnote{\url{https://imisglobal.com}}, our partner in the datAcron project, made available a dataset consisting of 55M position signals from 34K vessels sailing in the European seas between  January 1-30, 2016 \cite{datacron-del5.2, datacron-del5.3}. In contrast to the Brest dataset, signals transmitted via satellites are also included, thus capturing vessels sailing out of the range of terrestrial antennas. The synopsis generator labelled 17M position signals as critical, while the spatial preprocessing module produced 11M spatio-temporal events. In this dataset we did not have access to all areas of interest---e.g.~we did not have access to any anchorage areas away from the Brest area.
\vspace{0.1cm}

The experiments were conducted using the open-source RTEC composite event recognition engine\footnote{\url{https://github.com/aartikis/RTEC}}, under YAP-6 Prolog, in a machine running Ubuntu 16.04.3 LTS, with Intel Core i7-i7700 CPU @ 3.60GHz x 8 and 16 GB 2133 MHz RAM. The input of RTEC consisted of streams of the input events shown above the two horizontal lines of Table \ref{tbl:events}, while instances of all composite activities shown below these two lines were recognised.
For the Brest dataset, these composite activity instances are publicly available \cite{Mar-CER-dataset}.

\begin{table}[t]
\caption{Precision based on expert feedback.}
\label{tbl:pscores}\vspace{-0.4cm}
\renewcommand{\arraystretch}{0.9}
\setlength\tabcolsep{4pt}
\begin{tabular}{@{}lccc@{}}
\hline\noalign{\smallskip}
\textbf{Composite Event}                      & \textbf{TP}   & \textbf{FP}  & \textbf{Precision} \\ 
\noalign{\smallskip}
\hline
\noalign{\smallskip}
$\mathit{anchoredOrMoored(V)}$       & 3067 & 4   & 0.999     \\[2pt]
$\mathit{trawling(V)}$               & 29   & 0   & 1.000     \\[2pt]
$\mathit{tugging(V)}$                & 117  & 0   & 1.000     \\ [2pt]
$\mathit{pilotBoarding(V_1,V_2)}$         & 80   & 0   & 1.000     \\[2pt]
$\mathit{rendezVous(V_1,V_2)}$       & 52   & 2   & 0.963     \\[2pt]
\hline
\end{tabular}
\end{table}

\subsection{Accuracy}
\subsubsection{Expert Feedback}

The datasets presented above do not have an annotation of composite activities. 
For this reason, domain experts were asked to provide feedback on the recognised instances of a representative subset of the formalised activities: `anchored or moored', `trawling', `tugging', `pilot boarding' and `rendez-vous' \cite{cadets18}. 
Given that this evaluation process is highly time-consuming, as there are several thousand instances of recognised activities, the experts provided feedback only for the first month of the Brest dataset.
Table~\ref{tbl:pscores} presents the number of True Positives (TP), False Positives (FP) and the Precision score for the selected activities. (False Negatives and hence Recall could not be computed due to the absence of complete ground truth.)
The results show nearly perfect scores. The four False Positives of  $\mathit{anchoredOrMoored}$  were caused by four vessels that continued transmitting position signals while on land. Concerning $\mathit{rendezVous}$, the experts stated that for two recognised instances of this pattern, there were too few position signals to classify the activities as rendez-vous. Hence the two False Positives shown in Table \ref{tbl:pscores}.

\subsubsection{Compression Effects}

In addition to measuring accuracy based on expert feedback, we assessed the impact of trajectory compression on CER. We considered as reference the composite activities detected when consuming the complete AIS stream with the critical point labels---see the `enriched AIS stream' in Figure \ref{fig:prior-CER}---and then compared those activities against the ones detected when consuming the compressed AIS stream where all non-critical events have been removed---see the `critical point stream' in Figure \ref{fig:prior-CER}.

\begin{table}[t]
\centering
\small
\caption{Compression effects on accuracy.}
\label{tbl:recscores}\vspace{-0.4cm}
\renewcommand{\arraystretch}{0.9}
\setlength\tabcolsep{3pt}
\begin{tabular}{@{}lcccccc@{}}
\hline\noalign{\smallskip}
\multirow{2}{*}{\textbf{Composite Event}} & \multicolumn{6}{c}{\textbf{Brest} | \textbf{Europe}} \\
                            & \multicolumn{2}{c}{\textbf{Precision}} & \multicolumn{2}{c}{\textbf{Recall}} & \multicolumn{2}{c}{\textbf{$F_1$-Score}} \\ 
\noalign{\smallskip}
\hline
\noalign{\smallskip}
$\mathit{highSpeedNC(V)}$           & \multicolumn{1}{c|}{1.000} & 0.999  & \multicolumn{1}{c|}{0.978} & 0.979  & \multicolumn{1}{c|}{0.989} & 0.989     \\[2pt]
$\mathit{anchoredOrMoored(V)}$      & \multicolumn{1}{c|}{1.000} & 1.000  & \multicolumn{1}{c|}{1.000} & 1.000  & \multicolumn{1}{c|}{1.000} & 1.000     \\[2pt]
$\mathit{drifting(V)}$              & \multicolumn{1}{c|}{0.999} & -      & \multicolumn{1}{c|}{0.998} & -      & \multicolumn{1}{c|}{0.999} & -         \\[2pt]
$\mathit{trawling(V)}$              & \multicolumn{1}{c|}{0.988} & 0.997  & \multicolumn{1}{c|}{0.999} & 0.999  & \multicolumn{1}{c|}{0.994} & 0.998     \\[2pt]
$\mathit{tugging(V_1,V_2)}$         & \multicolumn{1}{c|}{0.991} & 0.968  & \multicolumn{1}{c|}{1.000} & 0.934  & \multicolumn{1}{c|}{0.994} & 0.951     \\[2pt]
$\mathit{pilotBoarding(V_1,V_2)}$   & \multicolumn{1}{c|}{1.000} & 1.000  & \multicolumn{1}{c|}{1.000} & 1.000  & \multicolumn{1}{c|}{1.000} & 1.000     \\[2pt]
$\mathit{rendezVous(V_1,V_2)}$      & \multicolumn{1}{c|}{1.000} & 1.000  & \multicolumn{1}{c|}{1.000} & 1.000  & \multicolumn{1}{c|}{1.000} & 1.000     \\[2pt]
$\mathit{loitering(V)}$             & \multicolumn{1}{c|}{1.000} & 1.000  & \multicolumn{1}{c|}{1.000} & 1.000  & \multicolumn{1}{c|}{1.000} & 1.000     \\[2pt]
$\mathit{sar(V)}$                   & \multicolumn{1}{c|}{0.998} & 0.987  & \multicolumn{1}{c|}{0.999} & 0.990  & \multicolumn{1}{c|}{0.998} & 0.988     \\[2pt]
\hline
\end{tabular}
\end{table}
\begin{figure}[t]
\centering
\includegraphics[width=0.9\columnwidth]{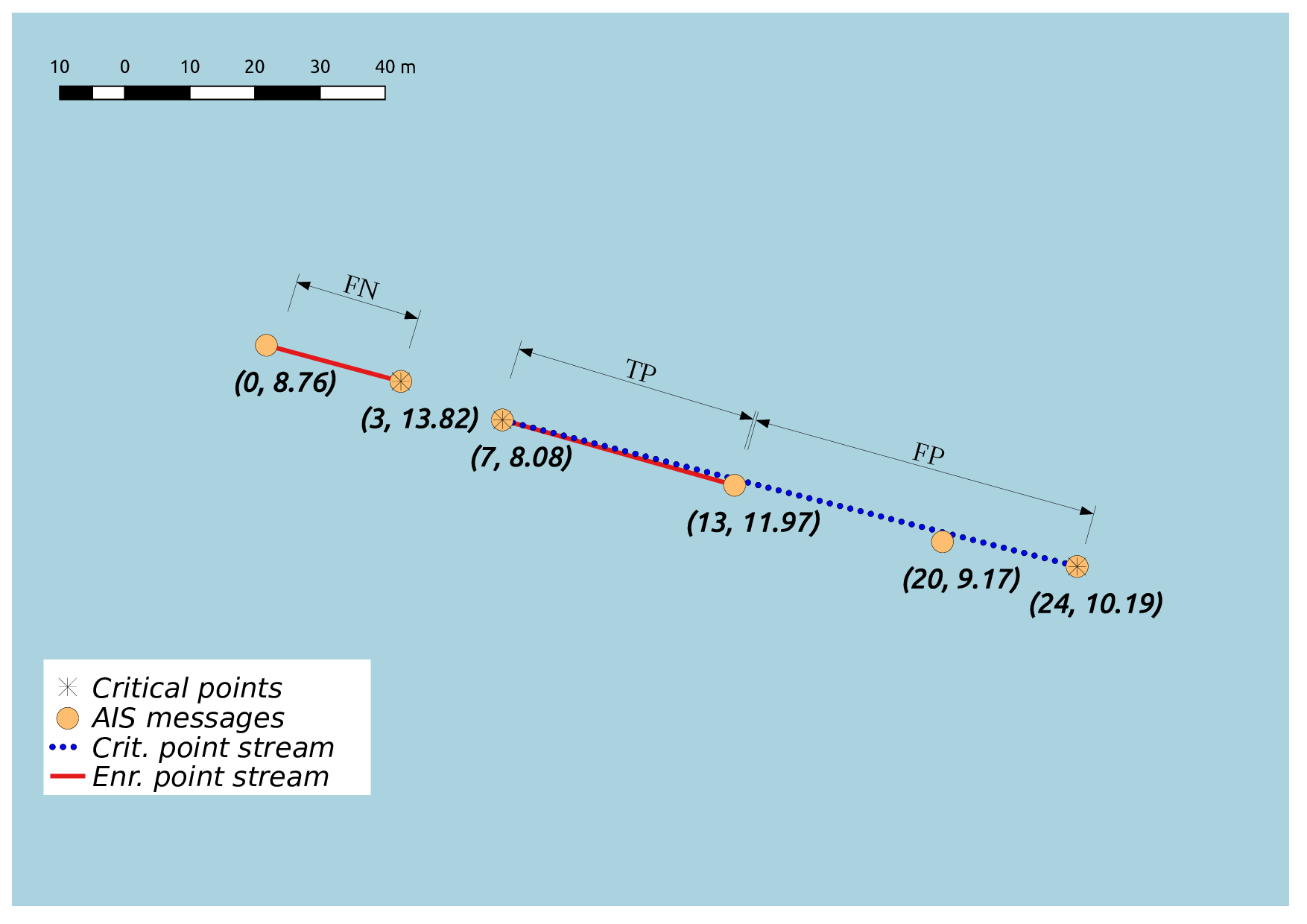}
\caption{Illustration of compression effects on CER using $\mathit{trawlingSpeed(V)}$. 
Red (respectively, blue dotted) lines denote the intervals during which $\mathit{trawlingSpeed(V)}\val\true$, as computed when consuming the enriched AIS (respectively, critical point) stream. 
The two numbers attached to each AIS message express the time-point of the message and the speed of the vessel.}
\label{fig:raw-crit}
\end{figure}

Table \ref{tbl:recscores} presents the comparison results in terms of Precision, Recall and $F_1$-score on the complete Brest dataset as well as the Europe dataset.
$\mathit{drifting}$ is missing from the latter dataset because information concerning the true heading of vessels was not available.
The set of True Positives expresses the time-points (seconds) in which a composite activity is recognised when consuming the enriched AIS stream and the critical point stream. 
Similarly, the set of False Positives (respectively, False Negatives) expresses the seconds in which a composite activity is recognised when consuming the critical point (respectively, enriched AIS) stream but not detected when consuming the enriched AIS (critical point) stream. 

\begin{figure*}[t]
\renewcommand{\thesubfigure}{a}
\subfloat[Input stream size.]{\centering
\includegraphics[width=.335\textwidth]{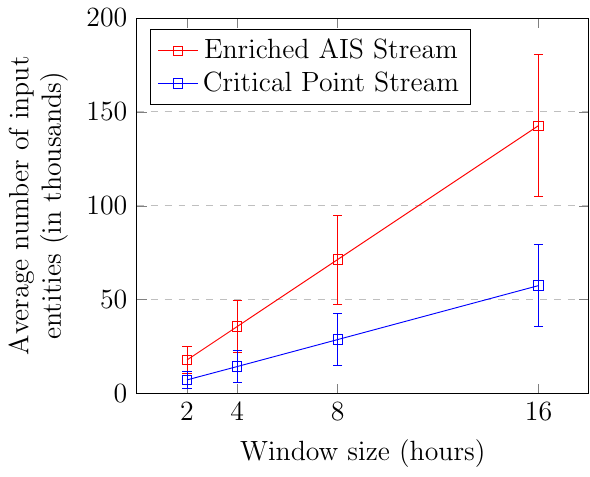}
\label{fig:Brest-IE}}
\renewcommand{\thesubfigure}{b}
\subfloat[Recognition times for all patterns.]{\centering
\includegraphics[width=.305\textwidth]{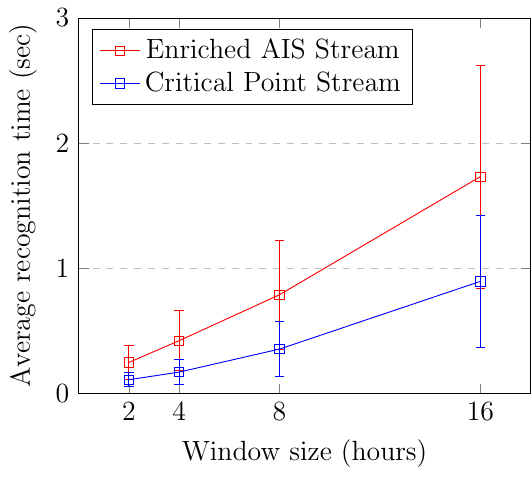}
\label{fig:Brest-rectimes}}
\renewcommand{\thesubfigure}{c}
\subfloat[Recognition times per pattern.]{\centering
\includegraphics[width=.31\textwidth]{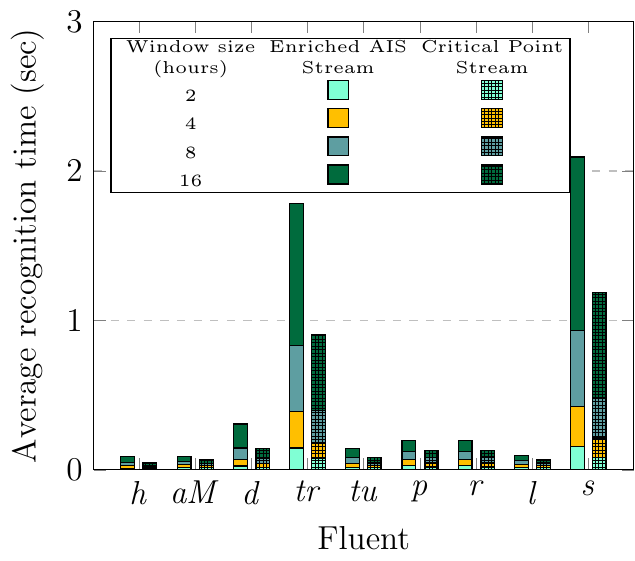}
\label{fig:Brest-rectimesperfluent}}\\

\renewcommand{\thesubfigure}{d}
\subfloat[Input stream size.]{\centering
\includegraphics[width=.32\textwidth]{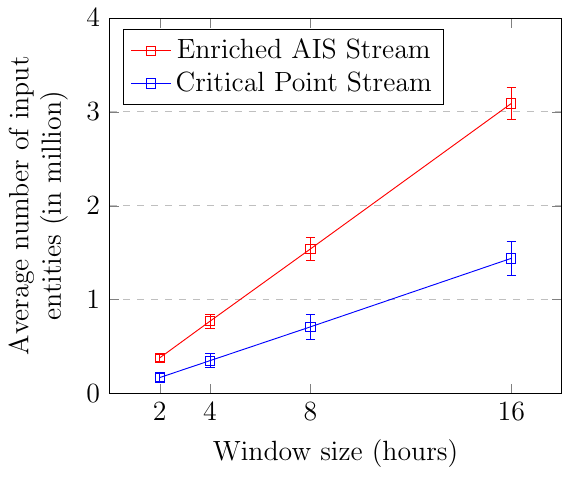}
\label{fig:EU-IE}}
\renewcommand{\thesubfigure}{e}
\subfloat[Recognition times for all patterns.]{\centering
\includegraphics[width=.31\textwidth]{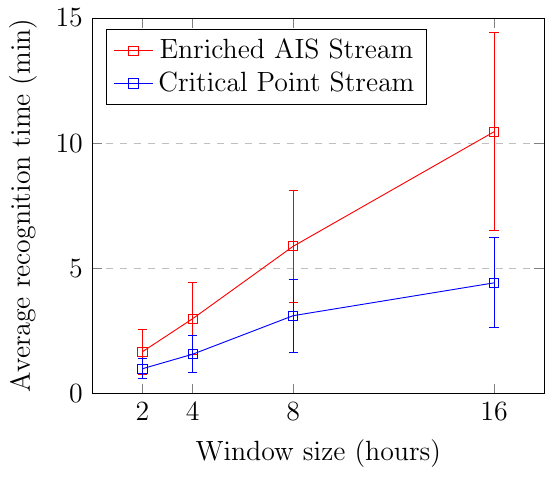}
\label{fig:EU-rectimes}}
\renewcommand{\thesubfigure}{f}
\subfloat[Recognition times per pattern.]{\centering
\includegraphics[width=.315\textwidth]{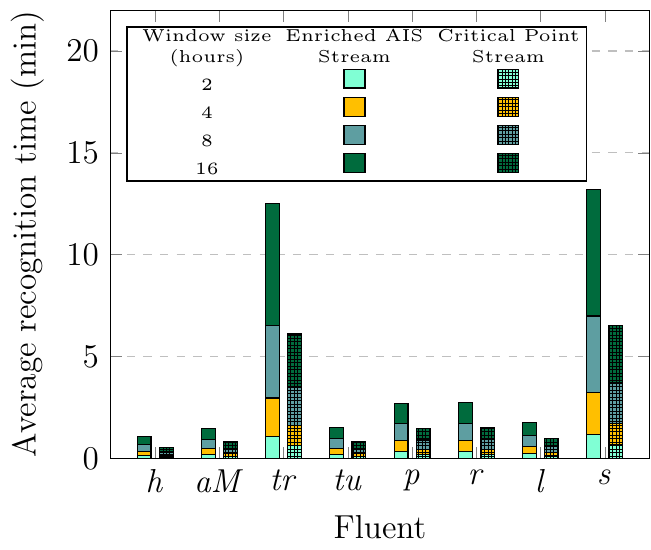}
\label{fig:EU-rectimesperfluent}}
\caption{CER Efficiency. Top diagrams: Brest dataset; bottom diagrams: Europe dataset. In the two right diagrams, $\mathit{h, aM, d, tr, tu, p, r, l, s}$ stand for $\mathit{highSpeedNC, anchoredOrMoored, drifting, trawling, tugging, pilotBoarding, rendezVous, loitering, sar}$.}
\label{fig:CER-efficiency}
\end{figure*}

Table \ref{tbl:recscores} shows that some of the recognised activities have perfect scores. The patterns of these activities are defined only in terms of critical points and spatio-temporal events, that are part of both the enriched AIS stream and the critical point stream.
For the remaining activities, there are some minor deviations between the results obtained by consuming the enriched AIS stream and the critical point stream. 
Those activities have some False Positives and False Negatives, thus compromising Precision and Recall. False Positives are created when a composite activity is terminated later when consuming the critical point stream than when consuming the enriched AIS stream. Similarly, False Negatives are created when an activity is initiated later when consuming the critical point stream than when consuming the enriched AIS stream. 

Figure~\ref{fig:raw-crit} illustrates this issue using $\mathit{trawlingSpeed(V)}$, which is used in the specification of trawling (see rule~\eqref{eq:trawling}). $\mathit{trawlingSpeed(V)}$ is initiated when the speed of the vessel in question is between 1--9 knots, and terminated otherwise (see Table~\ref{tbl:speed-related-bb}).
According to the example of Figure \ref{fig:raw-crit}, $\mathit{trawlingSpeed(V)}$ is initiated at time-point $\mathit{t\val 0}$ when the speed of the vessel is 8.76 knots, and terminated at $\mathit{t\val 3}$ when the speed becomes 13.82 knots. Similarly, $\mathit{trawlingSpeed}$ is initiated at $\mathit{t\val7}$ and terminated $\mathit{t\val13}$.
When consuming the critical point stream, however, different intervals are computed, since some of the AIS messages have been removed due to compression. For example, the AIS message at $t\val 0$ has been removed, thus not triggering the initiation of $\mathit{trawlingSpeed(V)}$ and consequently creating False Negatives. 
Moreover, the AIS message at $t\val 13$ has also been removed, delaying the termination of $\mathit{trawlingSpeed(V)}$, and thus creating False Positives.

\subsection{Efficiency}

Figure~\ref{fig:CER-efficiency} presents the experimental results in terms of efficiency on both datasets.
Results are shown for the enriched AIS stream and the critical point stream.
RTEC performs CER over a sliding window \cite{DBLP:journals/tkde/ArtikisSP15}. In these experiments, we varied the window size $\omega$ from 2 hours to 16 hours, while the slide step was set to 2 hours. Overlapping windows, as in the cases of $\omega\val 4, 8$ and $16$ hours in our experiments, are common in maritime monitoring, since AIS position signals may arrive with (substantial) delay at the CER system. 
This is especially the case for position signals arriving via satellites (i.e.~when vessels sail out of the range of terrestrial antennas), as in the case of the Europe dataset. 

Figures~\ref{fig:Brest-IE} and \ref{fig:EU-IE} display the average number of input events per window size, while
Figures~\ref{fig:Brest-rectimes} and \ref{fig:EU-rectimes} show the average recognition times for all patterns.
As expected, the performance is better in the smaller critical point stream rather than the enriched AIS stream. In both cases, RTEC performs real-time CER. For example, in the Brest dataset, a window of 16 hours, including more than 50K events in the critical point stream, is processed in less than a second in a single core of a standard desktop computer.

We may trivially parallelise CER by allocating different  patterns to different processing units. 
Figures~\ref{fig:Brest-rectimesperfluent} and \ref{fig:EU-rectimesperfluent} show the average recognition times per pattern. 
As discussed in Section \ref{sec:maritime}, the maritime patterns form a hierarchy, in the sense that the specification of a pattern depends on the specification of other, lower-level patterns. For example, the specification of $\mathit{tugging}$ depends on the specification of $\mathit{tuggingSpeed}$, $\mathit{gap}$ and $\mathit{withinArea}$ (see Figure \ref{fig:dependency-graph}). Thus, in Figures ~\ref{fig:Brest-rectimesperfluent} and \ref{fig:EU-rectimesperfluent} the displayed recognition times of $\mathit{tugging}$ include the recognition times of all patterns that contribute to its specification.

Figures~\ref{fig:Brest-rectimesperfluent} and \ref{fig:EU-rectimesperfluent} show that the most demanding patterns are those of $\mathit{trawling}$ and $\mathit{sar}$. This is due to the use of the `deadlines' mechanism of RTEC, whereby a fluent is automatically terminated some designated time after its last initiation.

\section{Summary and Further Work}
\label{sec:summary}

We presented a CER system for maritime monitoring extending our previous work \cite{DBLP:journals/geoinformatica/PatroumpasAAVPT17, Pitsikalis:2018:CEP:3200947.3201042}. 
The system includes a formal specification of effective maritime patterns that has been constructed in collaboration with, and evaluated by domain experts. In contrast to the related literature, these patterns concern a wide range of maritime activities. 
Furthermore, our evaluation on real datasets demonstrated that our system is capable of real-time CER for maritime monitoring.
The activities recognised on the 6-month dataset of the Brest area are publicly available \cite{Mar-CER-dataset}, in order to aid further research, such as the development of machine learning algorithms for pattern construction.

There are several directions for further work. We are implementing RTEC in Scala in order to pave the way for distributed CER. Additionally, we are developing machine learning techniques for continuously refining patterns given new data streaming into the system  \cite{DBLP:journals/tplp/KatzourisAP16,DBLP:journals/ml/KatzourisAP15}. Finally, in the context of the EU-funded INFORE project, we are integrating satellite images with position signals and geographical information for a more complete account of maritime situational awareness.

\begin{acks}

This work was supported by the datAcron project and the INFORE project, which have received funding from the European Union's Horizon 2020 research and innovation programme, under, respectively, grant agreement No 687591 and No 825070, and by the SYNTELESIS project (MIS 5002521), funded by the Operational Programme ``Competitiveness, Entrepreneurship and Innovation'' (NSRF 2014-2020), and co-financed by Greece and the European Union (European Regional Development Fund). We would like to thank Louis Le Masson and Jules Barre for their help in the evaluation of the maritime patterns.
Moreover, we would like to thank MarineTraffic\footnote{\url{https://www.marinetraffic.com/}} for giving us access to the vessel types for the dataset of Europe.

\end{acks}

\bibliographystyle{ACM-Reference-Format}
\bibliography{bibliography}

\end{document}